\documentclass[10pt,twocolumn,letterpaper]{article}

\usepackage{iccv}

\usepackage{times}
\usepackage{epsfig}
\usepackage{graphicx}
\usepackage{amsmath}
\usepackage{amssymb}
\usepackage{comment}
\usepackage{mathtools}
\usepackage{color}
\usepackage{subcaption}
\usepackage{url}            
\usepackage{booktabs}       
\usepackage{tabu}           
\usepackage{multirow}
\usepackage{algorithm}
\usepackage{algorithmicx}
\usepackage{algpseudocode}
\usepackage[dvipsnames]{xcolor}
\usepackage{color}
\usepackage{adjustbox}
\usepackage{colortbl}
\usepackage{pifont}

\iccvfinalcopy 

\newcommand{\tabincell}[2]{\begin{tabular}{@{}#1@{}}#2\end{tabular}}  

\definecolor{Gray}{gray}{0.95}

\makeatletter
\@namedef{ver@everyshi.sty}{}
\makeatother
\usepackage{tikz}
\usetikzlibrary{arrows,automata,positioning,shadows}

\newcommand{\paragrapha}[2]{\vspace{#1}\noindent\textbf{#2}}

\usepackage[pagebackref=true,breaklinks=true,colorlinks,bookmarks=false]{hyperref}

\def\iccvPaperID{7661} 

\ificcvfinal\fi

\begin{document}

\title{PoinTr: Diverse Point Cloud Completion with Geometry-Aware Transformers}
\author{Xumin Yu\thanks{Equal contribution.}, ~Yongming Rao\footnotemark[1], ~Ziyi Wang, ~Zuyan Liu, ~Jiwen Lu\thanks{Corresponding author.}, ~Jie Zhou\\
Department of Automation, Tsinghua University, China\\
State Key Lab of Intelligent Technologies and Systems, China\\
Beijing National Research Center for Information Science and Technology, China\\
{\tt\small yuxm20@mails.tsinghua.edu.cn; raoyongming95@gmail.com; } \\
{\tt\small \{wziyi20, liuzuyan19\}@mails.tsinghua.edu.cn; \tt\small \{lujiwen, jzhou\}@tsinghua.edu.cn} \\
}

\maketitle

\begin{abstract}

Point clouds captured in real-world applications are often incomplete due to the limited sensor resolution, single viewpoint, and occlusion. Therefore, recovering the complete point clouds from partial ones becomes an indispensable task in many practical applications. In this paper, we present a new method that reformulates point cloud completion as a set-to-set translation problem and design a new model, called PoinTr that adopts a transformer encoder-decoder architecture for point cloud completion. By representing the point cloud as a set of unordered groups of points with position embeddings, we convert the point cloud to a sequence of point proxies and employ the transformers for point cloud generation. To facilitate transformers to better leverage the inductive bias about 3D geometric structures of point clouds, we further devise a geometry-aware block that models the local geometric relationships explicitly. The migration of transformers enables our model to better learn structural knowledge and preserve detailed information for point cloud completion. Furthermore, we propose two more challenging benchmarks with more diverse incomplete point clouds that can better reflect the real-world scenarios to promote future research. Experimental results show that our method outperforms state-of-the-art methods by a large margin on both the new benchmarks and the existing ones. Code is available at \url{https://github.com/yuxumin/PoinTr}.
\end{abstract}

 \begin{figure}[!h]
  \centering
  \includegraphics[width = 0.98\linewidth]{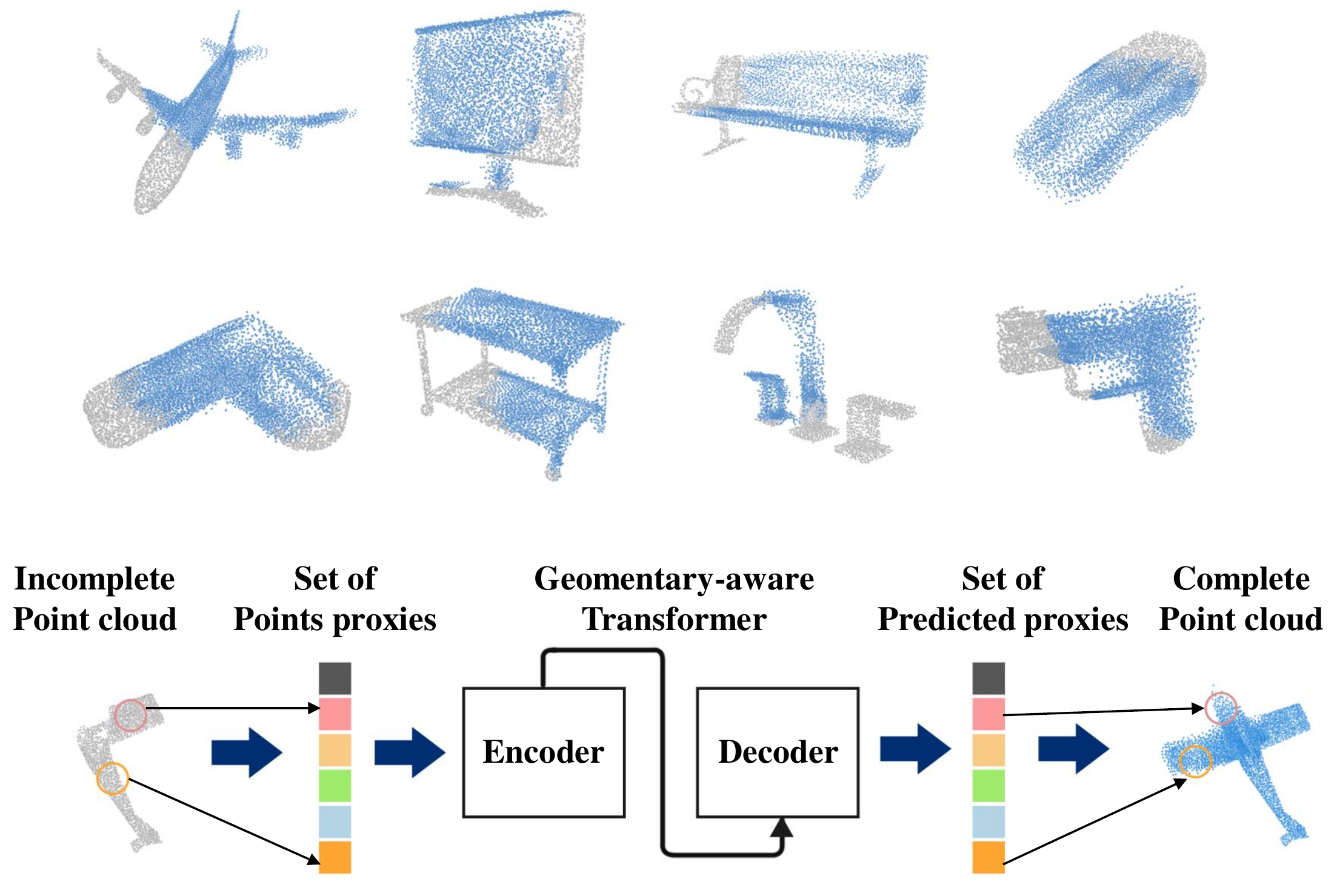}
  \caption{\small  \emph{PoinTr} is designed for point cloud completion task. It takes the downsampled partial point clouds as inputs ({\color{gray} gray points}), and predicts the missing parts and upsamples the known parts simultaneously ({\color{blue} blue points}). We propose to formulate the point cloud completion task as a set-to-set translation task and use a transformer encoder-decoder architecture to learn the complex dependencies among the point groups. Furthermore, we design two new benchmarks with more diverse tasks (\ie, upsampling and completion of point cloud), more diverse categories (\ie, from 8 categories to 55 categories), more diverse viewpoints (\ie, from 8 viewpoints to all possible viewpoints) and more diverse levels of incompleteness (\ie, missing 25\% to 75\% points of the ground-truth point clouds) to better reflect the real-world scenarios and promote future research. 
  }
  \label{fig:insight}
  \vspace{-10pt}
\end{figure}

\section{Introduction}
Recent developments in 3D sensors largely boost researches in 3D computer vision. One of the most commonly used 3D data format is the point cloud, which requires less memory to store but convey detailed 3D shape information. However, point cloud data from existing 3D sensors are not always complete and satisfactory because of inevitable self-occlusion, light reflection, limited sensor resolution, \etc. Therefore, recovering complete point clouds from partial and sparse raw data becomes an indispensable task with ever-growing significance. 

Over the years, researchers have tried many approaches to tackle this problem in the realm of deep learning. Early attempts on point cloud completion~\cite{DBLP:conf/cvpr/DaiQN17,DBLP:conf/iccv/HanLHKY17,DBLP:journals/corr/SharmaGF16,DBLP:conf/cvpr/StutzG18,DBLP:conf/cvpr/NguyenHTPY16,DBLP:conf/iros/VarleyDRRA17,DBLP:conf/nips/LiuTLH19,DBLP:conf/cvpr/LiuFXP19,DBLP:conf/cvpr/ZhouT18,yang20173d,wang2017shape} try to migrate mature methods from 2D completion tasks to 3D point clouds by voxelization and 3D convolutions. However, these methods suffer from a 
heavy computational cost that grows cubically as the spatial resolution increases. With the success of PointNet and PointNet++~\cite{DBLP:conf/cvpr/QiSMG17,DBLP:conf/nips/QiYSG17}, directly processing 3D coordinates becomes the mainstream of point cloud based 3D analysis. The technique is further applied to many pioneer works~\cite{DBLP:conf/iclr/AchlioptasDMG18,PCN,TopNet,PFNet,DBLP:journals/corr/abs-1901-08906,groueix2018papier,sarmad2019rl} in point cloud completion task, in which an encoder-decoder based architecture is designed to generate complete point clouds. However, the bottleneck of such methods lies in the max-pooling operation in the encoding phase, where fine-grained information is lost and can hardly be recovered in the decoding phase. 

Reconstructing complete point cloud is a challenging problem since the structural information required in the completion task runs counter to the unordered and unstructured nature of point cloud data. Therefore, learning structural features and long-range correlations among local parts of the point cloud becomes the key ingredient towards better point cloud completion. In this paper, we propose to adopt Transformers~\cite{Transformer}, one of the most successful architecture in Natural Language Processing (NLP), to learn the structural information of pairwise interactions and global correlations for point cloud completion. Our model, named \emph{PoinTr}, is characterized by five key components:  1) \emph{Encoder-Decoder Architecture}: We adopt the encoder-decoder architecture to convert point cloud completion as a set-to-set translation problem. The self-attention mechanism of transformers models all pairwise interactions between elements in the encoder, while the decoder reasons about the missing elements based on the learnable pairwise interactions among features of the input point cloud and queries; 2) \emph{Point Proxy}: We represent the set of point clouds in a local region as a feature vector called \emph{Point Proxy}. The input point cloud is convert to a sequence of Point Proxies, which are used as the inputs of our transformer model; 3) \emph{Geometry-aware Transformer Block}: To facilitate transformers to better leverage the inductive bias about 3D geometric structures of point clouds, we design a geometry-aware block that models the geometric relations explicitly; 4) \emph{Query Generator}: We use dynamic queries instead of fixed queirs in the decoder, which are generated by a query generation module that summarizes the features produced by the encoder and represents the initial sketch of the missing points; 5) \emph{Multi-Scale Point Cloud Generation}: We devise a multi-scale point generation module to recover the missing point cloud in a coarse-to-fine manner. 

As another contribution, we argue that existing benchmarks are not representative enough to cover real-world scenarios of incompleted point clouds. Therefore, we introduce two more challenging benchmarks that contain more diverse tasks (\ie, joint upsampling and completion of point cloud), more object categories (\ie, from 8 categories to 55 categories), more diverse views points (\ie, from 8 viewpoints to all possible viewpoints) and more diverse level of incompleteness (\ie, missing 25\% to 75\% points of the ground-truth point clouds). We evaluate our method on both the new benchmarks and the widely used PCN dataset~\cite{PCN} and KITTI benchmark~\cite{KITTI}. Experiments demonstrate that PointTr outperforms previous state-of-the-art methods on all benchmarks by a large margin. The main contributions of this paper are summarized in Figure~\ref{fig:insight}.


\section{Related Work}

 \begin{figure*}[th]
  \centering
  \includegraphics[width = \linewidth]{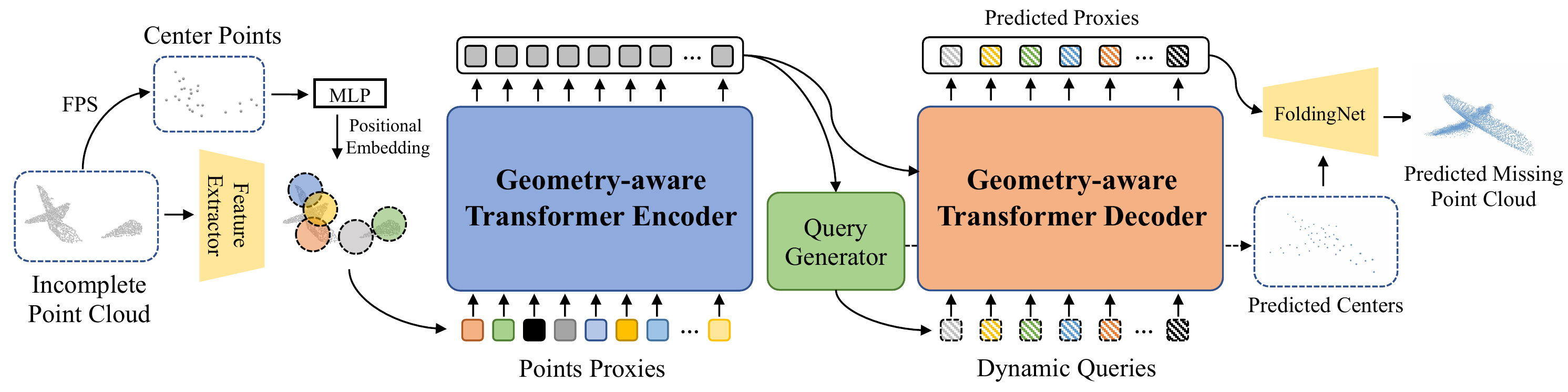}
  \caption{\small The Pipeline of \emph{PoinTr}. We first downsample the input partial point cloud to obtain the center points. Then, we use a lightweight DGCNN~\cite{wang2019DGCNN} to extract the local features around the center points. After adding the position embedding to the local feature, we use a transformer encoder-decoder architecture to predict the point proxies for the missing parts. A simple MLP and FoldingNet are used to complete the point cloud based on the predicted point proxies in a coarse-to-fine manner.}
  \label{fig:pipeline}
  \vspace{-10pt}
\end{figure*}

\label{sec:relatedwork}
\paragrapha{0pt}{3D Shape Completion. } 
Traditional methods for 3D shape completion tasks often adopt voxel grids or distance fields to describe 3D objects~\cite{DBLP:conf/cvpr/DaiQN17,DBLP:conf/iccv/HanLHKY17,DBLP:conf/cvpr/StutzG18}. Based on such structured 3D representations, the powerful 3D convolutions are used and achieve a great success in the tasks of 3D reconstruction~\cite{DBLP:conf/eccv/ChoyXGCS16,DBLP:conf/eccv/GirdharFRG16} and shape completion~\cite{DBLP:conf/cvpr/DaiQN17,DBLP:conf/iccv/HanLHKY17,DBLP:conf/cvpr/WuSKYZTX15}. However, this group of methods suffers from heavy memory consumption and computational burden. Although these issues are further alleviated by methods based on sparse representations~\cite{DBLP:conf/cvpr/SuJSMK0K18,DBLP:journals/tog/WangLGST17,DBLP:conf/cvpr/GrahamEM18}, the quantization operation in these methods still cause a significant loss in detailed information. Different from the above methods, researchers gradually start to use unstructured point clouds as the representation of 3D objects, given the small memory consumption and strong ability to represent fine-grained details. Nevertheless, the migration from structured 3D data understanding to point clouds analysis is non-trivial, since the commonly used convolution operator is no longer suitable for unordered points clouds. PointNet and its variants~\cite{DBLP:conf/cvpr/QiSMG17,DBLP:conf/nips/QiYSG17} are the pioneer work to directly process 3D coordinates and inspire the researches in many downstream tasks. In the realm of point cloud completion, PCN~\cite{PCN} is the first learning-based architecture, which proposes an Encoder-Decoder framework and adopts a FoldingNet to map the 2D points onto a 3D surface by mimicking the deformation of a 2D plane. After PCN, many other methods~\cite{TopNet,PFNet,GRNet,liu2020morphing} spring up, pursuing point clouds completion in higher resolution with better robustness.

\paragrapha{5pt}{Transformers. } Transformers~\cite{Transformer} are first introduced as an attention-based framework in Natural Language Processing (NLP). Transformer models often utilize the encoder-decoder architecture and are characterized by both self-attention and cross-attention mechanisms. Transformer models have proven to be very helpful to the tasks that involve long sequences thanks to the self-attention mechanism. The cross-attention mechanism in the decoder exploit the encoder information to learn the attention map of query features, which making transformers powerful in generation tasks. By taking the advantages of both self-attention and cross-attention mechanisms, transformers have a strong capability to handle long sequence input and enhance information communications between the encoder and the decoder. In the past few years, transformers have dominated the tasks that take long sequences as input and gradually replaced RNNs~\cite{vinyals2015order} in many domains. Now they begin their journey in computer vision~\cite{DBLP:conf/naacl/DevlinCLT19,DBLP:journals/corr/abs-1905-03072,2019arXiv191108460S,radford2019language,DBLP:journals/corr/abs-1802-05751,rao2021dynamicvit}.

\section{Approach}

The overall framework of \emph{PoinTr} is illustrated in Figure~\ref{fig:pipeline}. We will introduce our method in detail as follows.

\label{sec:method}

\subsection{Set-to-Set Translation with Transformers}

The primary goal of our method is to leverage the impressive sequence-to-sequence generation ability of transformer architecture for point cloud completion tasks. We propose to first convert the point cloud to a set of feature vectors, \textit{point proxies}, that represent the local regions in the point clouds (we will describe in Section~\ref{sec:proxy}). By analogy to the language translation pipeline, we model point cloud completion as a set-to-set translation task, where the transformers take the point proxies of the partial point clouds as the inputs and produce the point proxies of the missing parts. Specifically, given the set of point proxies $\mathcal{F} = \{F_1, F_2, ..., F_N\}$ that represents the partial point cloud, we model the process of point cloud completion as a set-to-set translation problem:
\begin{eqnarray} \small
    \mathcal{V} = \mathcal{M}_E(\mathcal{F}), \quad \mathcal{H} = \mathcal{M}_D( \mathcal{Q}, \mathcal{V}),
\end{eqnarray}
where $\mathcal{M}_E$ and $\mathcal{M}_D$ are the encoder and decoder models, $\mathcal{V} = \{V_1, V_2, ..., V_N\}$ are the output features of the encoder,  $\mathcal{Q} =  \{Q_1, Q_2, ..., Q_M\}$ are the dynamic queries  for the decoder, $\mathcal{H} =  \{H_1, H_2, ..., H_M\}$ are the predicted point proxies of the missing point cloud, and $M$ is the number of the predicted point proxies. The recent success in NLP tasks like text translation and question answering~\cite{devlin2018bert} have clearly demonstrated the effectiveness of transformers to solve this kind of problem. Therefore, we propose to adopt a transformer-based encoder-decoder architecture to solve the point cloud completion problem. 

The encoder-decoder architecture consists of $L_E$ and $L_D$ multi-head self-attention layers~\cite{Transformer} in the encoder and decoder, respectively. The self-attention layer in the encoder first updates proxy features with both long-range and short-range information. Then a feed forward network (FFN) further updates the proxy features with an MLP architecture. The decoder utilizes self-attention and cross-attention mechanisms to learn structural knowledge. The self-attention layer enhances the local features with global information, while the cross-attention layer explores the relationship between queries and outputs of the encoder. To predict the point proxies of the missing parts, we propose to use  dynamic query embeddings, which makes our decoder more flexible and adjustable for different types of objects and their missing information. More details about the transformer architecture can be found in the supplementary material and~\cite{devlin2018bert,Transformer}.

Note that benefiting from the self-attention mechanism in transformers, the features learned by the transformer network are invariant to the order of point proxies, which is also the basis of using transformers to process point clouds. Considering the strong ability to capture data relationships, we expect the transformer architecture to be a promising alternative for deep learning on point clouds. 

 \begin{figure}[t]
  \centering
  \includegraphics[width = \linewidth]{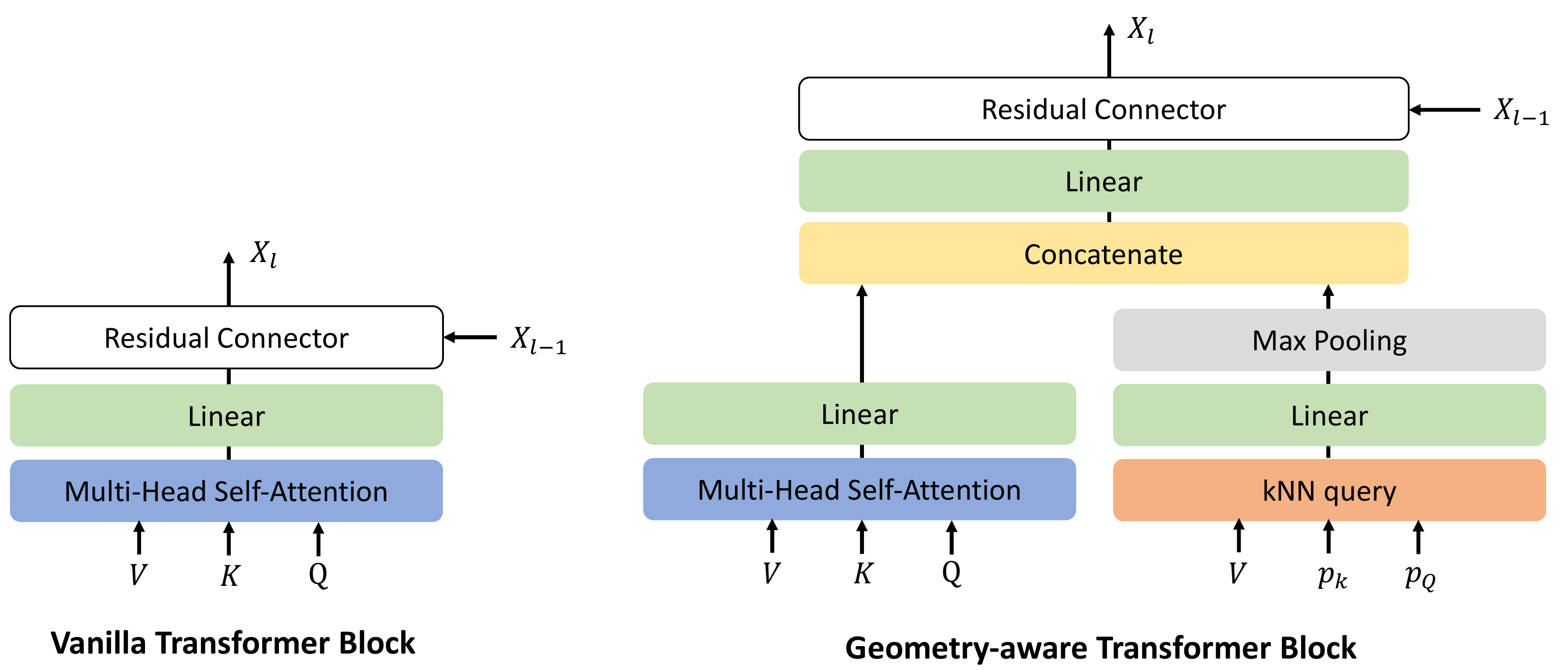}
  \caption{\small Comparisons of the vanilla transformer block and the proposed geometry-aware transformer block.}
  \label{fig:geo}
  \vspace{-10pt}
\end{figure}

\subsection{Point Proxy} \label{sec:proxy}

The Transformers in NLP take as input a 1D sequence of word  embeddings~\cite{Transformer}. To make 3D point clouds suitable for transformers, the first step is to convert the point cloud to a sequence of vectors. A trivial solution is directly feeding the sequence of $xyz$ coordinates to the transformers. However, since the computational complexity of the transformers is quadratic to the sequence length, this solution will lead to an unacceptable cost. Therefore, we propose to represent the original point cloud as a set of \textit{point proxies}. A point proxy represents a local region of the point clouds. Inspired by the set abstraction operation in~\cite{DBLP:conf/nips/QiYSG17}, we first conduct \textit{furthest point sample (FPS)} to locate a fixed number $N$ of point centers $\{q_1, q_2, ..., q_N\}$ in the partial point cloud. Then, we use a light-weight DGCNN~\cite{wang2019DGCNN} with hierarchical downsampling to extract the feature of the point centers from the input point cloud. The point proxy $F_i$ is a feature vector that captures the local structure around $q_i$, which can be computed as:
\begin{equation} \small
    F_i = F'_i + \varphi(q_i), \label{eq:proxy}
\end{equation}
where $F'_i$ is the feature of point $q_i$ that is extracted using the DGCNN model, and $\varphi$ is another MLP to capture the location information of the point proxy. The first term represents the semantic patterns of the local region, and the second term is inspired by the position embedding~\cite{bello2019attention} operation in transformers, which explicitly encodes the global location of the point proxy. The detailed architecture of the feature extraction model can be found in Supplementary Material.

\subsection{Geometry-aware Transformer Block} 

One of the key challenges of applying transformers for vision tasks is the self-attention mechanism in transformers lacks some inductive biases inherent to conventional vision models like CNNs and point cloud networks which explicitly model the structures of vision data. To facilitate transformers to better leverage the inductive bias about 3D geometric structures of point clouds, we design a geometry-aware block that models the geometric relations, which can be a plug-and-play module to incorporate with the attention blocks in any transformer architectures. The details of the proposed block are shown in Figure~\ref{fig:geo}. Different from the self-attention module that uses the feature similarity to capture the semantic relation, we propose to use kNN model to capture the geometric relation in the point cloud. Given the query coordinates $p_Q$, we query the features of the nearest keys according to the key coordinates $p_k$. Then we follow the practice of DGCNN~\cite{wang2019DGCNN} to learn the local geometric structures by feature aggregation with a linear layer followed by the max pooing operation. The geometric feature and semantic feature are then concatenated and mapped to the original dimensions to form the output.

\subsection{Query Generator}

The queries $\mathcal{Q}$ serve as the initial state of the predicted proxies. To make sure the queries correctly reflect the sketch of the completed point cloud, we propose a query generator module to generate the query embeddings dynamically conditioned on the encoder outputs. Specifically, we first summarize $\mathcal{V}$ with a linear projection to higher dimensions followed by the max pooing operation. Similar to~\cite{PCN}, we use a linear projection layer to directly generate $M\times3$ dimension features that can be reshaped as the $M$ coordinates $\{c_1, c_2, ..., c_M\}$. Lastly, we concatenate the global feature of the encoder and the coordinates, and use an MLP to produce the query embeddings. 

\subsection{Multi-Scale Point Cloud Generation}

The goal of our encoder-decoder network is to predict the missing parts of incomplete point clouds. However,  we can only get predictions for missing proxies from the transformer decoder. Therefore, we propose a multi-scale point cloud generation framework to recover missing point clouds at full resolution. To reduce redundant computations, we reuse the $M$ coordinates produced by the query generator as the local centers of the missing point cloud. Then, we utilize a FoldingNet~\cite{FoldingNet} $f$ to recover detailed local shapes centered at the predicted proxies:
\begin{equation} \small
    {\mathcal{P}}_i = f(H_i) + c_i, \quad i=1,2,...,M.
\end{equation}
where ${\mathcal{P}}_i$ is the set of neighboring points centered at $c_i$. Following previous work~\cite{PFNet}, we only predict the missing parts of the point cloud and concatenate them with the input point cloud to obtain the complete objects. Both predicted proxies and recovered point clouds are supervised during the training process, and the detailed loss function will be introduced in the following section.

\subsection{Optimization}\label{loss}

The loss function for point cloud completion should provide a quantitative measurement for the quality of output. However, since the point clouds are unordered, many loss functions that directly measure the distance between two points (\ie $\ell_2$ distance) are unsuitable. Fan \etal~\cite{fan2017point} introduce two metrics that are invariant to the permutation of points, which are Chamfer Distance (CD) and Earth Mover's Distance (EMD). We adopt Chamfer Distance as our loss function for its $\mathcal{O}(N\log N)$ complexity. We use $\mathcal{C}$ to represent the ${n_{\mathcal{C}}}$ local centers and  $\mathcal{P}$ to represent ${n_{\mathcal{P}}}$ points of the completed point cloud. Give the ground-truth completed point cloud $\mathcal{G}$, the loss functions for these two predictions can be written as:
\begin{small}
\begin{eqnarray} \small
\label{equ:Loss_func_2_1}
\nonumber
J_0&=& \frac{1}{n_{\mathcal{C}}}\sum_{c\in \mathcal{C}} \min_{g\in \mathcal{G}} \|c-g\| + \frac{1}{n_{\mathcal{G}}}\sum_{g\in \mathcal{G}} \min_{c\in \mathcal{C}} \|g-c\|,    \\   
\nonumber
J_1&=& \frac{1}{n_{\mathcal{P}}}\sum_{p\in \mathcal{P}} \min_{g\in \mathcal{G}} \|p-g\| + \frac{1}{n_{\mathcal{G}}}\sum_{g\in \mathcal{G}} \min_{p\in \mathcal{P}} \|g-p\|.
\end{eqnarray}
\end{small}
Note that we also concatenate the predicted local centers and the centers of the input point cloud to form the local centers of the whole object   $\mathcal{C}$. We directly use the high-resolution point cloud $\mathcal{G}$ to supervise the sparse point cloud $\mathcal{C}$ to encourage them to have similar distributions. Our final objective function is the sum of these two objectives $J = J_0 + J_1$.


\section{Experiments}

In this section, we first introduce the new benchmarks for diverse point cloud completion and the evaluation metric. Then, we show the results of both our method and several baseline methods on our new benchmarks. Lastly, we demonstrate the effectiveness of our model on the widely used PCN dataset and KITTI benchmark. We also provide ablation study and visual analysis of our method. 

\begin{table*}[t]
\small 
\caption{\small Results of our methods and state-of-the-art methods on ShapeNet-55. We report the detailed results for each method on 10 categories and the overall results on 55 categories for three difficulty degrees. We use CD-S, CD-M and CD-H to represent the CD results under the \emph{Simple}, \emph{Moderate} and \emph{Hard} settings. We also provide results under the F-Score@1\% metric. }
\label{tab:ShapeNet-55}
\newcolumntype{g}{>{\columncolor{Gray}}c}
\centering
\setlength{\tabcolsep}{1mm}{
\begin{tabular}{l | c c c c c| c c c c c |c c c |g g}
\toprule
&Table&Chair&Airplane&Car&Sofa&\tabincell{c}{Bird\\house}&Bag&Remote&\tabincell{c}{Key\\board}&Rocket& CD-S & CD-M & CD-H & CD-Avg & F1\\
\hline
FoldingNet~\cite{FoldingNet}& 2.53 & 2.81 & 1.43  & 1.98&2.48 &4.71&2.79&1.44&1.24&1.48&2.67&2.66&4.05&3.12& 0.082\\
PCN~\cite{PCN}& 2.13 & 2.29 & 1.02 & 1.85&2.06&4.50&2.86&1.33&0.89&1.32&1.94&1.96&4.08&2.66 & 0.133\\
TopNet~\cite{TopNet}& 2.21 & 2.53 & 1.14  & 2.18&2.36&4.83&2.93&1.49&0.95&1.32&2.26&2.16&4.3&2.91 & 0.126\\
PFNet~\cite{PFNet}& 3.95  & 4.24  & 1.81  & 2.53&3.34&6.21&4.96&2.91 &1.29&2.36&3.83&3.87&7.97&5.22 & 0.339\\
GRNet~\cite{GRNet}& 1.63 & 1.88 & 1.02  & 1.64&1.72 &2.97&2.06&1.09&0.89&1.03&1.35&1.71&2.85&1.97 & 0.238\\
\hline
PoinTr    &\textbf{0.81}  & \textbf{0.95}  & \textbf{0.44}  & \textbf{0.91} &\textbf{0.79}&\textbf{1.86}&\textbf{0.93}&\textbf{0.53}&\textbf{0.38}&\textbf{0.57}&\textbf{0.58}&\textbf{0.88}&\textbf{1.79}&\textbf{1.09} & \textbf{0.464}\\
\bottomrule
\end{tabular}}
\end{table*}

\begin{table*}[t] \small 
\caption{\small Results of our methods and state-of-the-art methods on ShapeNet-34. We report the results of 34 seen categories and 21 unseen categories in three difficulty degrees. We use CD-S, CD-M and CD-H to represent the CD results under the \emph{Simple}, \emph{Moderate} and \emph{Hard} settings.  We also provide results under the F-Score@1\% metric.
} 
\newcolumntype{g}{>{\columncolor{Gray}}c}
\label{tab:ShapeNet-34}
\centering
\setlength{\tabcolsep}{9pt}{
\begin{tabular}{lcccggcccgg}
\toprule
\multirow{2}[0]{*}{} & \multicolumn{5}{c}{\textbf{34 seen categories}} & \multicolumn{5}{c}{\textbf{21 unseen categories}} \\
\cmidrule(lr){2-6}\cmidrule(lr){7-11} 
& CD-S & CD-M & CD-H & CD-Avg & F1  & CD-S & CD-M & CD-H & CD-Avg & F1 \\
\hline
FoldingNet~\cite{FoldingNet}& 1.86 & 1.81 & 3.38  & 2.35 & 0.139 &2.76&2.74&5.36&3.62&0.095\\
PCN~\cite{PCN}&1.87 & 1.81 & 2.97 & 2.22 & 0.154 &3.17&3.08&5.29&3.85&0.101\\
TopNet~\cite{TopNet}& 1.77 & 1.61 & 3.54  & 2.31 & 0.171  &2.62&2.43&5.44&3.50&0.121\\
PFNet~\cite{PFNet}& 3.16  & 3.19  & 7.71  & 4.68& 0.347  &5.29&5.87&13.33&8.16&0.322\\
GRNet~\cite{GRNet}& 1.26 & 1.39 & 2.57  & 1.74  & 0.251 &1.85&2.25&4.87&2.99&0.216\\
\hline
 PoinTr    &  \textbf{0.76}  & \textbf{1.05}  & \textbf{1.88}  &\textbf{1.23} &\textbf{0.421} &\textbf{1.04}&\textbf{1.67}&\textbf{3.44}&\textbf{2.05}&\textbf{0.384}\\
\bottomrule
\end{tabular}}
\end{table*}

\label{sec:experiment}

\subsection{Benchmarks for Diverse Point Completion}
 We choose to generate the samples in our benchmarks based on the synthetic dataset, ShapeNet~\cite{ShapeNet}, because it contains the complete object models that cannot be obtained from real-world datasets like ScanNet~\cite{dai2017scannet} and S3DIS~\cite{armeni2017joint}. What makes our benchmarks distinct is that our benchmarks contain more object categories, more incomplete patterns and more viewpoints. Besides, we pay more attention to the ability of networks to deal with the objects from novel categories that do not appear in the training set.

\paragrapha{3pt}{ShapeNet-55 Benchmark:} In this benchmark, we use all the objects in ShapeNet from 55 categories.  Most existing datasets for point cloud completion like PCN~\cite{PCN} only consider a relatively small number of categories (\eg, 8 categories in PCN). However, the incompleted point clouds from the real-world scenarios are much more diverse. Therefore, we propose to evaluate the point cloud completion models on all 55 categories in ShapeNet to more comprehensively test the ability of models with a more diverse dataset.  We split the original ShapeNet using the 80-20 strategy: we randomly sample 80\% objects  from each category to form the training set and use the rest for evaluation. As a result, we get 41,952 models for training and 10,518 models for testing. For each object, we randomly sample 8,192 points from the surface to obtain the point cloud. 


\paragrapha{3pt}{ShapeNet-34 Benchmark:} In this benchmark, we want to explore another important issue in point cloud completion: the performance on novel categories. 
We believe it is necessary to build a benchmark for this task to better evaluate the generalization performance of models. We first split the origin ShapeNet into two parts: 21 unseen categories and 34 seen categories. In the seen categories, we randomly sample 100 objects from each category to construct a test set of the seen categories (3,400 objects in total) and leave the rest as the training set, resulting in 46,765 object models for training. We also construct another test set consisting of 2,305 objects from 21 novel categories. We evaluate the performance on both the seen and unseen categories to show the generalization ability of models.


\paragrapha{3pt}{Training and Evaluation:} In both benchmarks, the partial point clouds for training are generated online. We sample 2048 points from the object as the input and 8192 points as the ground truth. In order to mimic the real-world situation, we first randomly select a viewpoint and then remove the $n$ furthest points from the viewpoint to obtain a training partial point cloud. Although the projection method proposed in~\cite{PCN} is a better approximation to real scans, our strategy is more flexible and efficient. Our experiments on KITTI also show the model learned on our dataset works well when  finetuning to real-world scans. Besides, our strategy also ensures the diversity of our training samples in the aspect of viewpoints. During training, $n$ is randomly chosen from 2048 to 6144 (25\% to 75\% of the complete point cloud), resulting in different level of incompleteness. We then down-sample the remaining point clouds to 2048 points as the input data for models.

During evaluation, we fix 8 view points and $n$ is set to 2048, 4096 or 6144 (25\%, 50\% or 75\% of the whole point cloud) for convenience. According to the value of $n$, we divide the test samples into three difficulty degrees, \textit{simple}, \textit{moderate} and \textit{hard} in our experiments. In the following experiments, we will report the performance for each method in \textit{simple}, \textit{moderate} and \textit{hard} to show the ability of each network to deal with tasks at difficulty levels. In addition, we use the average performance under three difficulty degrees to report the overall performance (\textit{Avg}). 


\subsection{Evaluation Metric}
We follow the existing works~\cite{PCN,TopNet,PFNet,GRNet} to use the mean Chamfer Distance as the evaluation metric, which can measure distance between the prediction point cloud and ground-truth in set-level. For each prediction, the Chamfer Distance between the prediction point set $\mathcal{P}$ and the ground-truth point set $\mathcal{G}$ is calculated by:
\begin{small}
\begin{eqnarray} \centering
\nonumber
d_{CD}(\mathcal{P},\mathcal{G}) = \frac{1}{|\mathcal{P}|}\sum_{p\in \mathcal{P}} \min_{g\in \mathcal{G}} \|p-g\| + \frac{1}{|\mathcal{G}|}\sum_{g\in \mathcal{G}} \min_{p\in \mathcal{P}} \|g-p\|
\end{eqnarray}
\end{small}
Following the previous methods, we use two versions of Chamfer distance as the evaluation metric to compare the performance with existing works. CD-$\ell_1$ uses L1-norm to calculate the distance between two points, while CD-$\ell_2$ uses L2-norm instead. We also follow~\cite{FScore} to adopt F-Score as another evaluation metric. 

\subsection{Results on ShapeNet-55}

We first conduct experiments on ShapeNet-55, which consists of objects from 55 categories. To compare with existing methods, We implement FoldingNet~\cite{FoldingNet}, PCN~\cite{PCN}, TopNet~\cite{TopNet}, PFNet~\cite{PFNet} and GRNet~\cite{GRNet} on our benchmark according to their open-source code and use the best hyper-parameters in their papers for fair comparisons. We first investigate how the existing methods and our method perform when there are objects from more categories. The last four columns in Table~\ref{tab:ShapeNet-55} show that our PoinTr can better cope with different situations with diverse viewpoints, diverse object categories, diverse incomplete patterns and diverse incompleteness levels. We achieve 0.58, 0.6 and 0.69 improvement in CD-$\ell_2$ (multiplied by 1000) under three settings (simple, moderate and hard) comparing with the SOTA method GRNet~\cite{GRNet}. PFNet~\cite{PFNet}, which proposes to directly predict the missing parts of objects, fail in our benchmarks due to the high diversity. We further report the performance on categories with sufficient and insufficient samples. We only sample 10 categories out from 55 categories to show the results due to the limited space, in which \textit{Table}, \textit{chair}, \textit{Airplane},\textit{Car} and \textit{Sofa} contain more than 2500 samples in the training set while \textit{Birdhouse}, \textit{Bag}, \textit{Remote}, \text{Keyboard} and \textit{Rocket} contain less than 80 samples. We also provide the detailed results for all 55 categories in our supplementary material. We place the categories with sufficient samples at the first five columns and the categories with insufficient samples in the following five columns in Table~\ref{tab:ShapeNet-55}. The average CD results for three difficulty degrees are also reported. Surprisingly, there is no obvious difference between the results on these two kinds of categories. However, except for our PoinTr and SOTA method GRNet, the imbalance in the number of training samples lead to a relatively high CD in the categories with insufficient samples. Besides, our model achieves 0.46 F-Score on ShapeNet-55, while the state-of-the-art GRNet only obtain 0.24 F-Score. These results clearly demonstrate the effectiveness of PoinTr under the more diverse setting.

\subsection{Results on ShapeNet-34}
\begin{figure}[t]
  \centering
  \includegraphics[width = \linewidth]{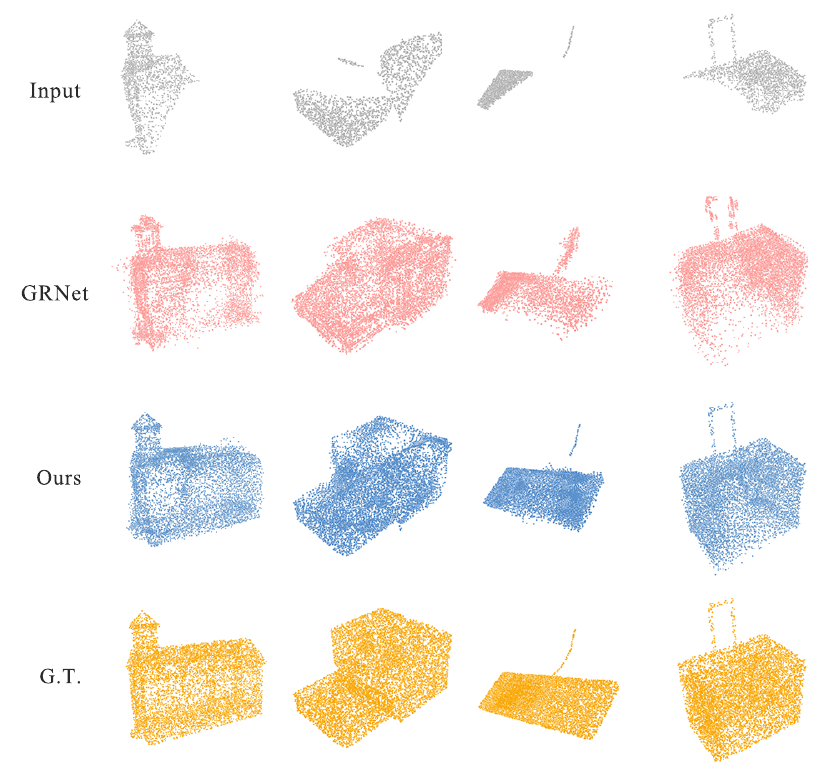}
  \caption{\small Point cloud completion results on some objects from novel categories. We show the input point cloud and the ground truth as well as the predictions of GRNet and our model. }
  \label{fig:novel}
\end{figure}

On ShapeNet-34, we also conduct experiments for our method and other five state-of-the-art methods. 
The results are shown in Table~\ref{tab:ShapeNet-34}. For the 34 seen categories, we can see our method outperforms all the other methods. For the 21 unseen categories, we using the networks that are trained on the 34 seen categories to evaluate the performance on the novel objects from the other 21 categories that do not appear in the training phase. We see our method also achieves the best performance in this more challenging setting. Comparing with the results of seen categories, we see in the simple setting (25\% of point cloud will be removed), the performance drop of our method is 
less than 0.3. But when the difficulty level increases, the performance gap between seen categories and unseen categories significantly increases. We also visualize the results in Figure~\ref{fig:novel} to show the effectiveness of our method on the unseen categories.

 \begin{figure}[t]
  \centering
  \includegraphics[width = \linewidth]{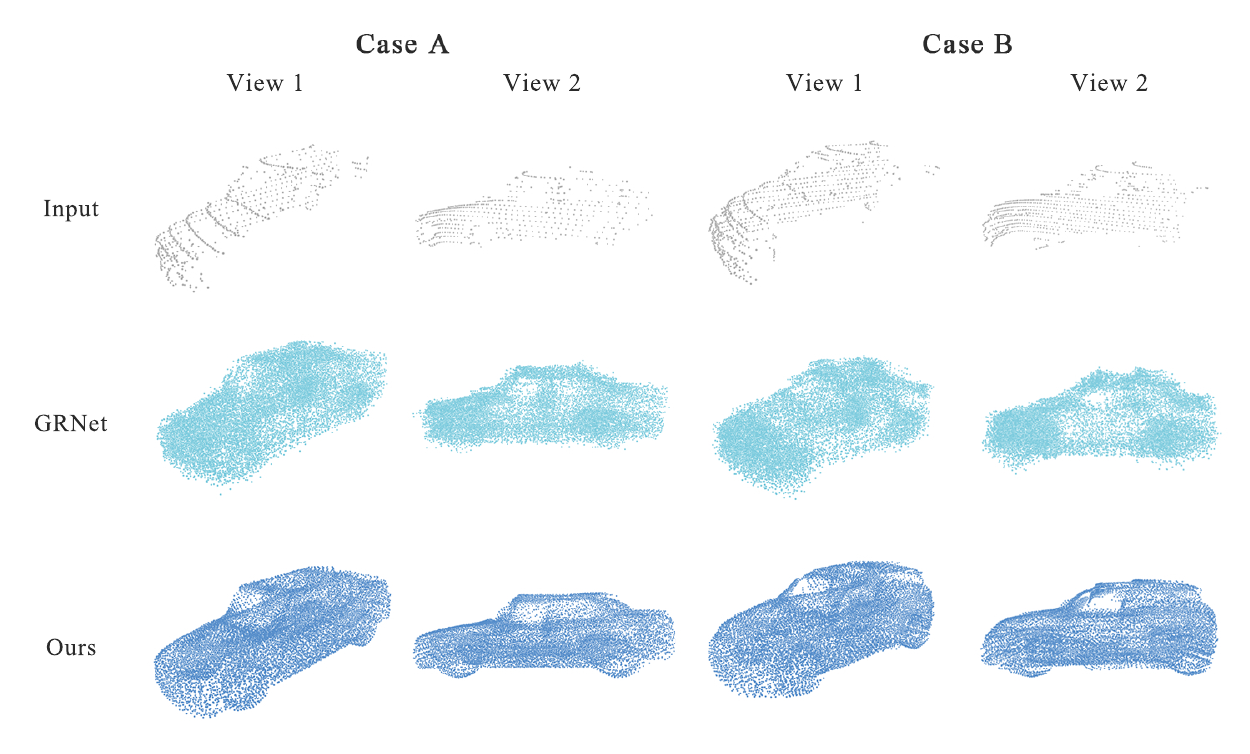}
  \caption{\small Qualitative results on the KITTI dataset. In order to better show the shape of the car, we provide two views of the same point cloud in each case. Our method can recover the complete point cloud of a car with more accurate boundaries and details (\eg tires of cars).}
  \label{fig:KITTI}
\end{figure}

\begin{table}[t]
\caption{\small Results on the PCN dataset. We use the L1 Chamfer Distance to compare with other methods.} 
\label{tab:PCN_benchmark}
\centering
\newcolumntype{g}{>{\columncolor{Gray}}c}
\setlength{\tabcolsep}{1.5mm}{
\begin{adjustbox}{width=\linewidth} \small
\begin{tabular}[\linewidth]{l | g | c c c c c c c c } 
\toprule
CD-$\ell_1$ ($\times$ 1000) & Avg & Air & Cab & Car & Cha & Lam & Sof & Tab & Wat  \\
\hline
FoldingNet~\cite{FoldingNet}& 14.31 & 9.49 & 15.80 & 12.61 & 15.55 & 16.41 & 15.97 & 13.65 & 14.99\\
AtlasNet~\cite{AtlasNet}& 10.85& 6.37& 11.94 &10.10& 12.06& 12.37 &12.99& 10.33 &10.61\\
PCN~\cite{PCN}& 9.64 & 5.50 & 22.70 & 10.63 & 8.70 & 11.00 & 11.34 & 11.68 & 8.59\\
TopNet~\cite{TopNet}& 12.15 & 7.61&  13.31&  10.90&  13.82&  14.44&  14.78&  11.22&  11.12\\
MSN~\cite{MSN}&10.0 &5.6& 11.9& 10.3 &10.2&10.7& 11.6&9.6&9.9 \\
GRNet~\cite{GRNet}& 8.83 &6.45& 10.37 &9.45 &9.41& 7.96 &10.51 &8.44 &8.04\\
PMP-Net~\cite{PMP}&8.73& 5.65 &11.24& 9.64& 9.51 &6.95 &10.83& 8.72& 7.25\\
CRN~\cite{CRN} &8.51& 4.79&9.97& 8.31& 9.49& 8.94& 10.69& 7.81& 8.05\\
\hline
PoinTr &\textbf{8.38}  &4.75&10.47&8.68&9.39&7.75&10.93&7.78&7.29\\
\bottomrule[1pt]
\end{tabular}
\end{adjustbox}}
\end{table}

\begin{table}[t]
\small
\caption{\small Ablation study on the PCN dataset. We investigate different designs including query generator (Query), DGCNN feature extractor (DGCNN) and Geometry-aware Blocks (Geometry). } 
\label{tab:Ablation Study}
\centering
\newcolumntype{g}{>{\columncolor{Gray}}c}
\setlength{\tabcolsep}{1.5mm}{
\begin{adjustbox}{width=0.45\textwidth}
\begin{tabular}[\linewidth]{c | c c c |gg }

\toprule
Model  & Query & DGCNN & Geometry & CD-$\ell_1$ & F-Score@1\%\\
\midrule
A & & & & 9.43 & 67.82 \\
B &\checkmark & & & 9.09 & 0.713 \\
C    &\checkmark &\checkmark & & 8.69 & 0.736 \\
D     &\checkmark &\checkmark & all & 8.44 & 0.741 \\
E    &\checkmark &\checkmark & $1^{\rm st}$ & \textbf{8.38} & \textbf{0.745} \\
\bottomrule
\end{tabular}
\end{adjustbox}}
\end{table}

\begin{table*}[t]
\caption{Results on LiDAR scans from KITTI dataset under the  Fidelity and MMD metrics.} 
\vspace{-5pt}
\label{tab:KITTI}
\centering
\newcolumntype{g}{>{\columncolor{Gray}}c}
\begin{adjustbox}{width=\textwidth}
\begin{tabular}[\linewidth]{l | c c c c c c c c c | g}
\toprule
CD-$\ell_2$ ($\times$ 1000) & AtlasNet~\cite{AtlasNet} & PCN~\cite{PCN} & FoldingNet~\cite{FoldingNet} & TopNet~\cite{TopNet} & MSN~\cite{MSN} & NSFA~\cite{zhang2020detail} & PFNet~\cite{PFNet} &   CRN~\cite{CRN} & GRNet~\cite{GRNet} & PoinTr\\
\midrule
Fidelity $\downarrow$ & 1.759 & 2.235 & 7.467 &  5.354 &  0.434 & 1.281 & 1.137  & 1.023 & 0.816 & \textbf{0.000} \\
MMD $\downarrow$ & 2.108 & 1.366 & 0.537 & 0.636 & 2.259 &0.891 & 0.792& 0.872& 0.568  & \textbf{0.526} \\
\bottomrule
\end{tabular}
\end{adjustbox}
\end{table*}

 \begin{figure}[t]
  \centering
  \includegraphics[width = \linewidth]{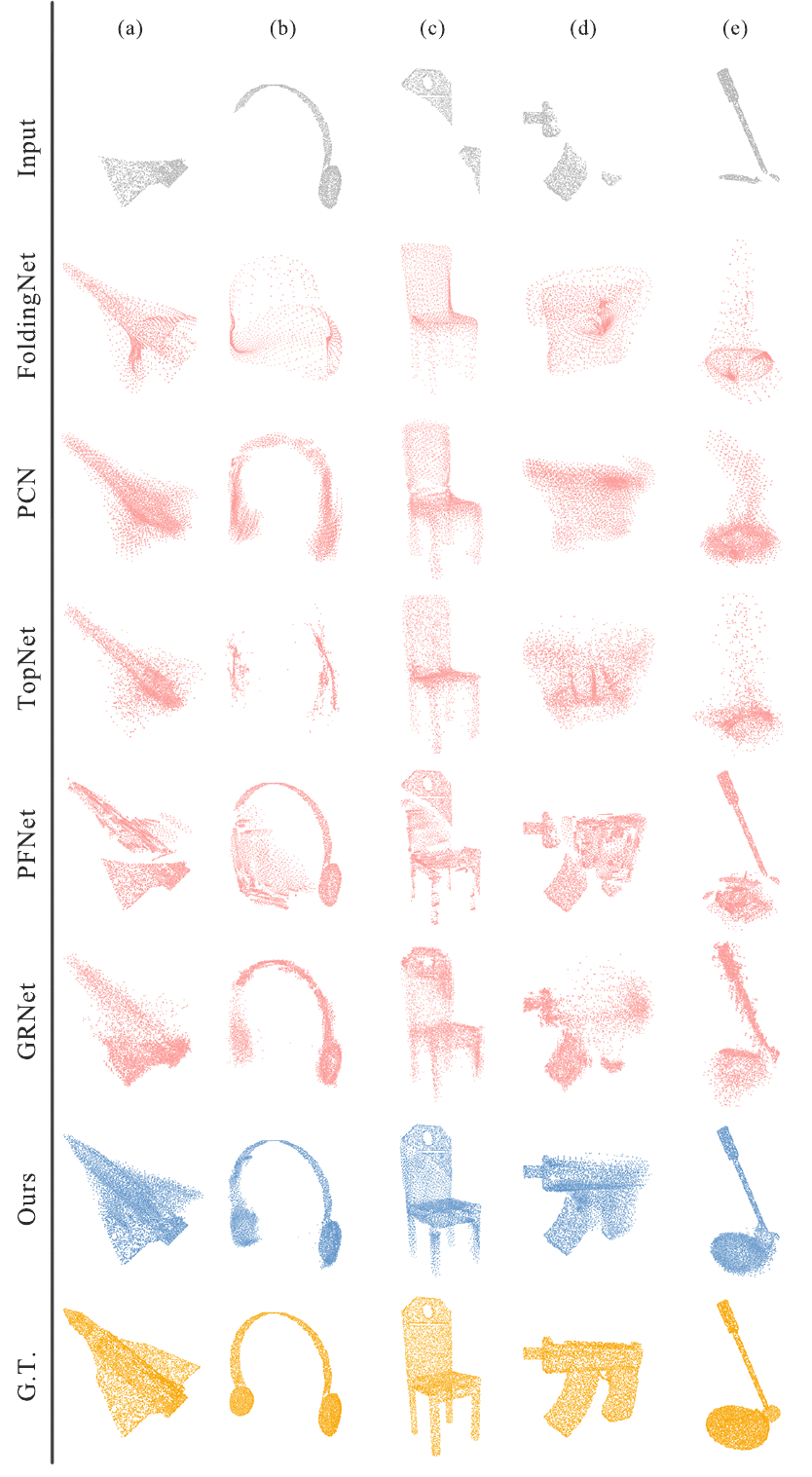}
  \caption{ Qualitative results on ShapeNet-55. All methods above takes the point clouds in the first line as inputs and generate complete point clouds. Our methods can complete the point clouds with higher fidelity, which clearly shows the effectiveness of our method.} \vspace{-10pt}
  \label{fig:case}
\end{figure}

\subsection{Results on the Existing Benchmarks}
Apart from the experiments on the two newly proposed challenging benchmarks, we also conduct the experiments on the existing benchmarks including the PCN dataset~\cite{PCN} and KITTI benchmark~\cite{KITTI}.

\paragrapha{5pt}{Results on the PCN Dataset. } The PCN dataset~\cite{PCN} is one of the most widely used benchmark datasets for the point cloud completion task. To verify the effectiveness of our method on existing benchmarks and compare it with more state-of-the-art methods, we conducted experiments on this dataset following the standard protocol and evaluation metric used in previous work~\cite{PCN,MSN,GRNet,PMP,CRN}. The results are shown in Table~\ref{tab:PCN_benchmark}. We see our method largely improves the previous methods and establishes the new state-of-the-art on this dataset.

\paragrapha{5pt}{Results on KITTI Benchmark. }
To show the performance of our method in real-world scenarios, we follow~\cite{GRNet} to finetune our trained model on ShapeNetCars~\cite{PCN} and evaluate the performance of our model on KITTI dataset, which contains the incomplete point clouds of cars in the real-world scenes from LiDAR scans. We report the Fidelity and MMD metrics in Table~\ref{tab:KITTI} and show some reconstruction results in Figure~\ref{fig:KITTI}. Our method achieves better qualitative and quantitative performance.

\subsection{Model Design Analysis}
To examine the effectiveness of our designs, we conduct a detailed ablation study on the key components of PoinTr. The results are summarized in
Table~\ref{tab:Ablation Study}. The baseline model A is the vanilla transformer model for point cloud completion, which uses the encoder-decoder architecture with the standard transformer blocks. In this model, we form the point proxies directly from the point cloud using a single-layer DGCNN model. We then add the query generator between the encoder and decoder (model B). We see the query generator improve the baseline by 0.34 in Chamfer distance. When using DGCNN to extract features from the input point cloud (model C), we observe a significant improvement to 8.69. By adding the geometric block to all the transformer blocks (model D), we see the performance can be further improved, which clearly demonstrates the effectiveness of the geometric structures learned by the block. We find that only adding the geometric block to the first transformer block in both encoder and decoder can lead to a slightly better performance (model E), which indicates the role of geometric block is to introduce the inductive bias and a single layer is sufficient while adding more blocks may result in over-fitting. Besides, we see our method can achieve over 0.74 F-Score on the PCN dataset while obtaining only 0.46 F-Score on our ShapeNet-55, which also suggests our new datatset is much more challenging.

\subsection{Qualitative Results}
\label{sec:quan}
In Figure~\ref{fig:case}, we show some completion results for all methods and find our method perform better. For example, the input data in (a) nearly lose all the geometric information and can be hardly recognized as an airplane. In this case, other methods can only roughly complete the shape with unsatisfactory geometry details, while our method can still complete the point cloud with higher fidelity. These results show our method has a stronger ability to recover details and is more robust to various incomplete patterns. More results can be found in the supplementary material. 


\section{Conclusion}
\label{sec:con}
In this paper, we have proposed a new architecture, PoinTr, to convert the point cloud completion task into a set to set translation tasks. With several technical innovations, we successfully applied the transformer model to this task and achieved state-of-the-art performance. Moreover, we proposed two more challenging benchmarks for more diverse point cloud completion. 
Extending our transformer architecture to other 3D tasks can an interesting future direction.

\subsection*{Acknowledgements}
This work was supported in part by the National Key Research and Development Program of China under Grant 2017YFA0700802, in part by the National Natural Science Foundation of China under Grant 61822603, Grant U1813218, and Grant U1713214, in part by a grant from the Beijing Academy of Artificial Intelligence (BAAI), and in part by a grant from the Institute for Guo Qiang, Tsinghua University.

\begin{appendix}
\section{Implementation Details}

Our proposed method PointTr is implemented with PyTorch~\cite{paszke2019pytorch}. We utilize AdamW optimizer~\cite{loshchilov2018fixing} to train the network with initial learning rate as 0.0005 and weight decay as 0.0005. In all of our experiments, we set the depth of the encoder and decoder in our transformer to 6 and 8 and set $k$ of kNN operation to 16 and 8 for the DGCNN feature extractor and the geometry-aware block respectively. We use 6 head attention for all transformer blocks and set their hidden dimensions to 384. On the PCN dataset, the network takes 2048 points as inputs and is required to complete the other 14336 points. We set the batch size to 54 and train the model for 300 epochs with the continuous learning rate decay of 0.9 for every 20 epochs. We set $N$ to 128 and $M$ to 224. On ShapeNet-55/34, the model takes 2048 points as inputs and is required to complete the other 6144 points. We set the batch size to 128 and train the model for 200 epochs with the continuous learning rate decay of 0.76 for every 20 epochs. We set $N$ to 128 and $M$ to 96.

We employ a lightweight DGCNN~\cite{wang2019DGCNN} model to extract the point proxy features.  To reduce the computational cost, we hierarchically downsample the original input point cloud to $N=128$ center points and use several DGCNN layers to capture local geometric relationships. The detailed network architecture is: $\texttt{Linear}(C_{in}=3, C_{out}=8)$ $\rightarrow$ $\texttt{DGCNN}(C_{in}=8, C_{out}=32, K=8, N_{out}=2048)$ $\rightarrow$ $\texttt{DGCNN}(C_{in}=32, C_{out}=64, K=8, N_{out}=512)$ $\rightarrow$ $\texttt{DGCNN}(C_{in}=64, C_{out}=64, K=8, N_{out}=512)$ $\rightarrow$ $\texttt{DGCNN}(C_{in}=64, C_{out}=128, K=8, N_{out}=128)$, where $C_{in}$ and $C_{out}$ are the numbers of channels of input and output features, $N_{out}$ is the number of points after FPS.

\section{Technical Details on Transformers}

\begin{figure}[t]
  \centering
  \includegraphics[width = 0.8\linewidth]{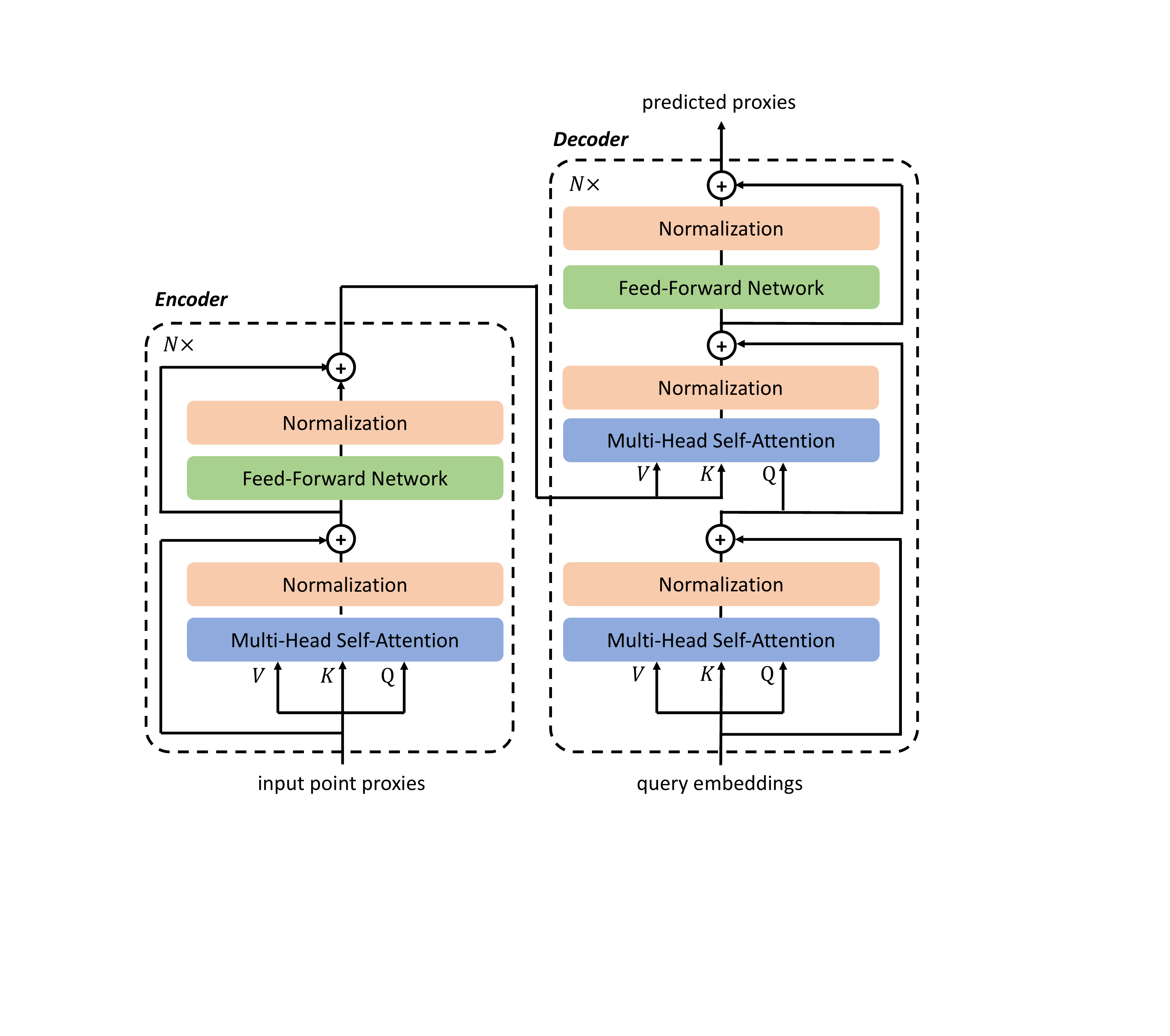}
  \caption{\small The overall architecture of the transformer encoder-decoder networks.}
  \label{fig:trans}
\end{figure}

\noindent \textbf{Encoder-Decoder Architecture. } The overall architecture of the transformer encoder-decoder networks is illustrated in Figure~\ref{fig:trans}. The point proxies are passed through the transformer encoder with $N$ multi-head self-attention layers and feed-forward network layers. Then, the decoder receives the generated query embeddings and encoder memory, and produces the final set of predicted point proxies that represents the missing part of the point cloud through $N$ multi-head self-attention layers, decoder-encoder attention layers and feed-forward network layers. We set $N$ to 6 in all our experiments following common practice~\cite{Transformer}.

\vspace{5pt}\noindent \textbf{Multi-head Attention. } Multi-head attention mechanism allows the network to jointly attend to information from different representation subspaces at different positions~\cite{Transformer}. Speciacally, given the input values $V$, keys $K$ and queries $Q$, the multi-head attention is computed by:
\begin{equation}
    \begin{split}
        \text{MultiHead}(Q, K, V) = W^O \text{Concat}(\text{head}_1, ..., \text{head}_h), \nonumber
    \end{split}
\end{equation}
where $W^O$ the weights of the output linear layer and each head feature can be obtained by:
\begin{equation}
    \begin{split}
        \text{head}_i = \text{softmax}(\frac{QW_i^Q (KW_i^K)^T}{\sqrt{d_k}}) VW_i^V \nonumber
    \end{split}
\end{equation}
where $W_i^Q$, $W_i^K$ and $W_i^V$ are the linear layers that project the inputs to different subspaces and $d_k$ is the dimension of the input features. 

\vspace{5pt} \noindent \textbf{Feed-forward network (FFN). } Following~\cite{Transformer}, we use two linear layers with ReLU
activations and dropout as the feed-forward network.

\section{Detailed Experimental Results }

\paragraph{Detailed results on ShapeNet-55:} 
In Table~\ref{Table:ShapeNet-55}, we report the detailed results for FoldingNet~\cite{FoldingNet}, PCN~\cite{PCN}, TopNet~\cite{TopNet}, PFNet~\cite{PFNet}, GRNet~\cite{GRNet} and the proposed method on ShapeNet-55. Each row in the table stands for a category of object. We test each method under three settings: simple, moderate and hard. 

\begin{table*}
\caption{
Detailed results on ShapeNet-55. \textit{S.}, \textit{M.} and \textit{H.} stand for the simple, moderate and hard settings.}
\label{Table:ShapeNet-55} 
\centering
\vspace{4pt}
\begin{adjustbox}{width=\textwidth}
\begin{tabular}[\linewidth]{l | c c c| c c c| c c c| c c c| c c c| c c c }
\toprule
\multirow{2}{*}{CD-$\ell_2$($\times$ 1000)}& \multicolumn{3}{c|}{FoldingNet~\cite{FoldingNet}}&\multicolumn{3}{c|}{PCN~\cite{PCN}}&\multicolumn{3}{c|}{TopNet~\cite{TopNet}}&\multicolumn{3}{c|}{PFNet~\cite{PFNet}}&\multicolumn{3}{c|}{GRNet~\cite{GRNet}}&\multicolumn{3}{c}{Ours-PoinTr}\\
\cline{2-19}
&S.&M.&H.&S.&M.&H.&S.&M.&H.&S.&M.&H.&S.&M.&H.&S.&M.&H.\\
\hline
airplane& 1.36 & 1.28 & 1.7 & 0.9 & 0.89 & 1.32 & 1.02 & 0.99 & 1.48 & 1.35 & 1.44 & 2.69 & 0.87 & 0.87 & 1.27 & \textbf{0.27} & \textbf{0.38} & \textbf{0.69}\\ 
trash bin& 2.93 & 2.9 & 5.03 & 2.16 & 2.18 & 5.15 & 2.51 & 2.32 & 5.03 & 4.03 & 3.39 & 9.63 & 1.69 & 2.01 & 3.48 & \textbf{0.8} & \textbf{1.15} & \textbf{2.15}\\ 
bag& 2.31 & 2.38 & 3.67 & 2.11 & 2.04 & 4.44 & 2.36 & 2.23 & 4.21 & 3.63 & 3.66 & 7.6 & 1.41 & 1.7 & 2.97 & \textbf{0.53} & \textbf{0.74} & \textbf{1.51}\\ 
basket& 2.98 & 2.77 & 4.8 & 2.21 & 2.1 & 4.55 & 2.62 & 2.43 & 5.71 & 4.74 & 3.88 & 8.47 & 1.65 & 1.84 & 3.15 & \textbf{0.73} & \textbf{0.88} & \textbf{1.82}\\ 
bathtub& 2.68 & 2.66 & 4.0 & 2.11 & 2.09 & 3.94 & 2.49 & 2.25 & 4.33 & 3.64 & 3.5 & 5.74 & 1.46 & 1.73 & 2.73 & \textbf{0.64} & \textbf{0.94} & \textbf{1.68}\\ 
bed& 4.24 & 4.08 & 5.65 & 2.86 & 3.07 & 5.54 & 3.13 & 3.1 & 5.71 & 4.44 & 5.36 & 9.14 & 1.64 & 2.03 & 3.7 & \textbf{0.76} & \textbf{1.1} & \textbf{2.26}\\ 
bench& 1.94 & 1.77 & 2.36 & 1.31 & 1.24 & 2.14 & 1.56 & 1.39 & 2.4 & 2.17 & 2.16 & 4.11 & 1.03 & 1.09 & 1.71 & \textbf{0.38} & \textbf{0.52} & \textbf{0.94}\\ 
birdhouse& 4.06 & 4.18 & 5.88 & 3.29 & 3.53 & 6.69 & 3.73 & 3.98 & 6.8 & 3.96 & 5.0 & 9.66 & 1.87 & 2.4 & 4.71 & \textbf{0.98} & \textbf{1.49} & \textbf{3.13}\\ 
bookshelf& 3.04 & 3.03 & 3.91 & 2.7 & 2.7 & 4.61 & 3.11 & 2.87 & 4.87 & 3.19 & 3.47 & 5.72 & 1.42 & 1.71 & 2.78 & \textbf{0.71} & \textbf{1.06} & \textbf{1.93}\\ 
bottle& 1.7 & 1.91 & 4.02 & 1.25 & 1.43 & 4.61 & 1.56 & 1.66 & 4.02 & 2.37 & 2.89 & 10.03 & 1.05 & 1.44 & 2.67 & \textbf{0.37} & \textbf{0.74} & \textbf{1.5}\\ 
bowl& 2.79 & 2.6 & 4.23 & 2.05 & 1.83 & 3.66 & 2.33 & 1.98 & 4.82 & 4.3 & 3.97 & 8.76 & 1.6 & 1.77 & 2.99 & \textbf{0.68} & \textbf{0.78} & \textbf{1.44}\\ 
bus& 1.47 & 1.42 & 2.0 & 1.2 & 1.14 & 2.08 & 1.32 & 1.21 & 2.29 & 2.06 & 1.88 & 3.75 & 1.06 & 1.16 & 1.48 & \textbf{0.42} & \textbf{0.55} & \textbf{0.79}\\ 
cabinet& 2.0 & 1.86 & 2.79 & 1.6 & 1.49 & 3.47 & 1.91 & 1.65 & 3.36 & 2.72 & 2.37 & 4.73 & 1.27 & 1.41 & 2.09 & \textbf{0.55} & \textbf{0.66} & \textbf{1.16}\\ 
camera& 5.5 & 6.04 & 8.87 & 4.05 & 4.54 & 8.27 & 4.75 & 4.98 & 9.24 & 6.57 & 8.04 & 13.11 & 2.14 & 3.15 & 6.09 & \textbf{1.1} & \textbf{2.03} & \textbf{4.34}\\ 
can& 2.84 & 2.68 & 5.71 & 2.02 & 2.28 & 6.48 & 2.67 & 2.4 & 5.5 & 5.65 & 4.05 & 16.29 & 1.58 & 2.11 & 3.81 & \textbf{0.68} & \textbf{1.19} & \textbf{2.14}\\ 
cap& 4.1 & 4.04 & 5.87 & 1.82 & 1.76 & 4.2 & 3.0 & 2.69 & 5.59 & 10.92 & 9.04 & 20.3 & 1.17 & 1.37 & 3.05 & \textbf{0.46} & \textbf{0.62} & \textbf{1.64}\\ 
car& 1.81 & 1.81 & 2.31 & 1.48 & 1.47 & 2.6 & 1.71 & 1.65 & 3.17 & 2.06 & 2.1 & 3.43 & 1.29 & 1.48 & 2.14 & \textbf{0.64} & \textbf{0.86} & \textbf{1.25}\\ 
cellphone& 1.04 & 1.06 & 1.87 & 0.8 & 0.79 & 1.71 & 1.01 & 0.96 & 1.8 & 1.25 & 1.37 & 3.65 & 0.82 & 0.91 & 1.18 & \textbf{0.32} & \textbf{0.39} & \textbf{0.6}\\ 
chair& 2.37 & 2.46 & 3.62 & 1.7 & 1.81 & 3.34 & 1.97 & 2.04 & 3.59 & 2.94 & 3.48 & 6.34 & 1.24 & 1.56 & 2.73 & \textbf{0.49} & \textbf{0.74} & \textbf{1.63}\\ 
clock& 2.56 & 2.41 & 3.46 & 2.1 & 2.01 & 3.98 & 2.48 & 2.16 & 4.03 & 3.15 & 3.27 & 6.03 & 1.46 & 1.66 & 2.67 & \textbf{0.62} & \textbf{0.84} & \textbf{1.65}\\ 
keyboard& 1.21 & 1.18 & 1.32 & 0.82 & 0.82 & 1.04 & 0.88 & 0.83 & 1.15 & 0.83 & 1.06 & 1.97 & 0.74 & 0.81 & 1.09 & \textbf{0.3} & \textbf{0.39} & \textbf{0.45}\\ 
dishwasher& 2.6 & 2.17 & 3.5 & 1.93 & 1.66 & 4.39 & 2.43 & 1.74 & 4.64 & 4.57 & 3.23 & 6.39 & 1.43 & 1.59 & 2.53 & \textbf{0.55} & \textbf{0.69} & \textbf{1.42}\\ 
display& 2.15 & 2.24 & 3.25 & 1.56 & 1.66 & 3.26 & 1.84 & 1.85 & 3.48 & 2.27 & 2.83 & 5.52 & 1.13 & 1.38 & 2.29 & \textbf{0.48} & \textbf{0.67} & \textbf{1.33}\\ 
earphone& 6.37 & 6.48 & 9.14 & 3.13 & 2.94 & 7.56 & 4.36 & 4.47 & 8.36 & 15.07 & 17.5 & 33.37 & 1.78 & 2.18 & 5.33 & \textbf{0.81} & \textbf{1.38} & \textbf{3.78}\\ 
faucet& 4.46 & 4.39 & 7.2 & 3.21 & 3.48 & 7.52 & 3.61 & 3.59 & 7.25 & 5.68 & 6.79 & 14.29 & 1.81 & 2.32 & 4.91 & \textbf{0.71} & \textbf{1.42} & \textbf{3.49}\\ 
filecabinet& 2.59 & 2.48 & 3.76 & 2.02 & 1.97 & 4.14 & 2.41 & 2.12 & 4.12 & 3.72 & 3.57 & 7.13 & 1.46 & 1.71 & 2.89 & \textbf{0.63} & \textbf{0.84} & \textbf{1.69}\\ 
guitar& 0.65 & 0.6 & 1.25 & 0.42 & 0.38 & 1.23 & 0.57 & 0.47 & 1.42 & 0.74 & 0.89 & 5.41 & 0.44 & 0.48 & 0.76 & \textbf{0.14} & \textbf{0.21} & \textbf{0.42}\\ 
helmet& 5.39 & 5.37 & 7.96 & 3.76 & 4.18 & 7.53 & 4.36 & 4.55 & 7.73 & 9.55 & 8.41 & 15.44 & 2.33 & 3.18 & 6.03 & \textbf{0.99} & \textbf{1.93} & \textbf{4.22}\\ 
jar& 3.65 & 3.87 & 6.51 & 2.57 & 2.82 & 6.0 & 3.03 & 3.17 & 7.03 & 5.44 & 5.56 & 11.87 & 1.72 & 2.37 & 4.37 & \textbf{0.77} & \textbf{1.33} & \textbf{2.87}\\ 
knife& 1.29 & 0.87 & 1.21 & 0.94 & 0.62 & 1.37 & 0.84 & 0.68 & 1.44 & 2.11 & 1.53 & 3.89 & 0.72 & 0.66 & 0.96 & \textbf{0.2} & \textbf{0.33} & \textbf{0.56}\\ 
lamp& 3.93 & 4.23 & 6.87 & 3.1 & 3.45 & 7.02 & 3.03 & 3.39 & 8.15 & 6.82 & 7.61 & 14.22 & 1.68 & 2.43 & 5.17 & \textbf{0.64} & \textbf{1.4} & \textbf{3.58}\\ 
laptop& 1.02 & 1.04 & 1.96 & 0.75 & 0.79 & 1.59 & 0.8 & 0.85 & 1.66 & 1.04 & 1.21 & 2.46 & 0.83 & 0.87 & 1.28 & \textbf{0.32} & \textbf{0.34} & \textbf{0.6}\\ 
loudspeaker& 3.21 & 3.15 & 4.55 & 2.5 & 2.45 & 5.08 & 3.1 & 2.76 & 5.32 & 4.32 & 4.19 & 7.6 & 1.75 & 2.08 & 3.45 & \textbf{0.78} & \textbf{1.16} & \textbf{2.17}\\ 
mailbox& 2.44 & 2.61 & 4.98 & 1.66 & 1.74 & 5.18 & 2.16 & 2.1 & 5.1 & 3.82 & 4.2 & 10.51 & 1.15 & 1.59 & 3.42 & \textbf{0.39} & \textbf{0.78} & \textbf{2.56}\\ 
microphone& 4.42 & 5.06 & 7.04 & 3.44 & 3.9 & 8.52 & 2.83 & 3.49 & 6.87 & 6.58 & 7.56 & 16.74 & 2.09 & 2.76 & 5.7 & \textbf{0.7} & \textbf{1.66} & \textbf{4.48}\\ 
microwaves& 2.67 & 2.48 & 4.43 & 2.2 & 2.01 & 4.65 & 2.65 & 2.15 & 5.07 & 4.63 & 3.94 & 6.52 & 1.51 & 1.72 & 2.76 & \textbf{0.67} & \textbf{0.83} & \textbf{1.82}\\ 
motorbike& 2.63 & 2.55 & 3.52 & 2.03 & 2.01 & 3.13 & 2.29 & 2.25 & 3.54 & 2.17 & 2.48 & 5.09 & 1.38 & 1.52 & 2.26 & \textbf{0.75} & \textbf{1.1} & \textbf{1.92}\\ 
mug& 3.66 & 3.67 & 5.7 & 2.45 & 2.48 & 5.17 & 2.89 & 2.56 & 5.43 & 4.76 & 4.3 & 8.37 & 1.75 & 2.16 & 3.79 & \textbf{0.91} & \textbf{1.17} & \textbf{2.35}\\ 
piano& 3.86 & 4.04 & 6.04 & 2.64 & 2.74 & 4.83 & 2.99 & 2.89 & 5.64 & 4.57 & 5.26 & 9.26 & 1.53 & 1.82 & 3.21 & \textbf{0.76} & \textbf{1.06} & \textbf{2.23}\\ 
pillow& 2.33 & 2.38 & 3.87 & 1.85 & 1.81 & 3.68 & 2.31 & 2.26 & 4.19 & 4.21 & 3.82 & 7.89 & 1.42 & 1.67 & 3.04 & \textbf{0.61} & \textbf{0.82} & \textbf{1.56}\\ 
pistol& 1.92 & 1.62 & 2.52 & 1.25 & 1.17 & 2.65 & 1.5 & 1.3 & 2.62 & 2.27 & 2.09 & 7.2 & 1.11 & 1.06 & 1.76 & \textbf{0.43} & \textbf{0.66} & \textbf{1.3}\\ 
flowerpot& 4.53 & 4.68 & 6.46 & 3.32 & 3.39 & 6.04 & 3.61 & 3.45 & 6.28 & 4.83 & 5.51 & 10.68 & 2.02 & 2.48 & 4.19 & \textbf{1.01} & \textbf{1.51} & \textbf{2.77}\\ 
printer& 3.66 & 4.01 & 5.34 & 2.9 & 3.19 & 5.84 & 3.04 & 3.19 & 5.84 & 5.56 & 6.06 & 9.29 & 1.56 & 2.38 & 4.24 & \textbf{0.73} & \textbf{1.21} & \textbf{2.47}\\ 
remote& 1.14 & 1.2 & 1.98 & 0.99 & 0.97 & 2.04 & 1.14 & 1.17 & 2.16 & 1.74 & 2.37 & 4.61 & 0.89 & 1.05 & 1.29 & \textbf{0.36} & \textbf{0.53} & \textbf{0.71}\\ 
rifle& 1.27 & 1.02 & 1.37 & 0.98 & 0.8 & 1.31 & 0.98 & 0.86 & 1.46 & 1.72 & 1.45 & 3.02 & 0.83 & 0.77 & 1.16 & \textbf{0.3} & \textbf{0.45} & \textbf{0.79}\\ 
rocket& 1.37 & 1.18 & 1.88 & 1.05 & 1.04 & 1.87 & 1.04 & 1.0 & 1.93 & 1.65 & 1.61 & 3.82 & 0.78 & 0.92 & 1.44 & \textbf{0.23} & \textbf{0.48} & \textbf{0.99}\\ 
skateboard& 1.58 & 1.58 & 2.07 & 1.04 & 0.94 & 1.68 & 1.08 & 1.05 & 1.84 & 1.43 & 1.6 & 3.09 & 0.82 & 0.87 & 1.24 & \textbf{0.28} & \textbf{0.38} & \textbf{0.62}\\ 
sofa& 2.22 & 2.09 & 3.14 & 1.65 & 1.61 & 2.92 & 1.93 & 1.76 & 3.39 & 2.65 & 2.53 & 4.84 & 1.35 & 1.45 & 2.32 & \textbf{0.56} & \textbf{0.67} & \textbf{1.14}\\ 
stove& 2.69 & 2.63 & 3.99 & 2.07 & 2.02 & 4.72 & 2.44 & 2.16 & 4.84 & 4.03 & 3.71 & 7.15 & 1.46 & 1.72 & 3.22 & \textbf{0.63} & \textbf{0.92} & \textbf{1.73}\\ 
table& 2.23 & 2.15 & 3.21 & 1.56 & 1.5 & 3.36 & 1.78 & 1.65 & 3.21 & 3.03 & 3.11 & 5.74 & 1.15 & 1.33 & 2.33 & \textbf{0.46} & \textbf{0.64} & \textbf{1.31}\\ 
telephone& 1.07 & 1.06 & 1.75 & 0.8 & 0.8 & 1.67 & 1.02 & 0.95 & 1.78 & 1.3 & 1.47 & 3.37 & 0.81 & 0.89 & 1.18 & \textbf{0.31} & \textbf{0.38} & \textbf{0.59}\\ 
tower& 2.46 & 2.45 & 3.91 & 1.91 & 1.97 & 4.47 & 2.15 & 2.05 & 4.51 & 3.13 & 3.54 & 9.87 & 1.26 & 1.69 & 3.06 & \textbf{0.55} & \textbf{0.9} & \textbf{1.95}\\ 
train& 1.86 & 1.68 & 2.32 & 1.5 & 1.41 & 2.37 & 1.59 & 1.44 & 2.51 & 2.01 & 2.03 & 4.1 & 1.09 & 1.14 & 1.61 & \textbf{0.5} & \textbf{0.7} & \textbf{1.12}\\ 
watercraft& 1.85 & 1.69 & 2.49 & 1.46 & 1.39 & 2.4 & 1.53 & 1.42 & 2.67 & 2.1 & 2.13 & 4.58 & 1.09 & 1.12 & 1.65 & \textbf{0.41} & \textbf{0.62} & \textbf{1.07}\\ 
washer& 3.47 & 3.2 & 4.89 & 2.42 & 2.31 & 6.08 & 2.92 & 2.53 & 6.53 & 5.55 & 4.11 & 7.04 & 1.72 & 2.05 & 4.19 & \textbf{0.75} & \textbf{1.06} & \textbf{2.44}\\ 
\hline
mean& 2.68 & 2.66 & 4.06 & 1.96 & 1.98 & 4.09 & 2.26 & 2.17 & 4.31 & 3.84 & 3.88 & 8.03 & 1.35 & 1.63 & 2.86 & \textbf{0.58} & \textbf{0.88} & \textbf{1.8}\\

\bottomrule
\end{tabular}
\end{adjustbox}
\vspace{20pt}
\end{table*}

\paragraph{Detailed results on ShapeNet-34: } 
In Table~\ref{Table:ShapeNet-21}, we report the detailed results for the novel objects from 21 categories in ShapeNet-34. Each row in the table stands for a category of object. We test each method under the three settings: simple, moderate and hard.

\begin{table*}
\caption{
Detailed results for the novel objects on ShapeNet-34. \textit{S.}, \textit{M.} and \textit{H.} stand for the simple, moderate and hard settings.}
\vspace{4pt}
\label{Table:ShapeNet-21} \centering
\begin{adjustbox}{width=\textwidth}
\begin{tabular}[\linewidth]{l | c c c| c c c| c c c| c c c| c c c| c c c }
\toprule
\multirow{2}{*}{CD-$\ell_2$ ($\times$ 1000)}& \multicolumn{3}{c|}{FoldingNet~\cite{FoldingNet}}&\multicolumn{3}{c|}{PCN~\cite{PCN}}&\multicolumn{3}{c|}{TopNet~\cite{TopNet}}&\multicolumn{3}{c|}{PFNet~\cite{PFNet}}&\multicolumn{3}{c|}{GRNet~\cite{GRNet}}&\multicolumn{3}{c}{Ours-PoinTr}\\
\cline{2-19}
&S.&M.&H.&S.&M.&H.&S.&M.&H.&S.&M.&H.&S.&M.&H.&S.&M.&H.\\
\hline
bag& 2.15 & 2.27 & 3.99 & 2.48 & 2.46 & 3.94 & 2.08 & 1.95 & 4.36 & 3.88 & 4.42 & 9.67 & 1.47 & 1.88 & 3.45 & \textbf{0.96} & \textbf{1.34} & \textbf{2.08}\\ 
basket& 2.37 & 2.2 & 4.87 & 2.79 & 2.51 & 4.78 & 2.46 & 2.11 & 5.18 & 4.47 & 4.55 & 14.46 & 1.78 & 1.94 & 4.18 & \textbf{1.04} & \textbf{1.4} & \textbf{2.9}\\ 
birdhouse& 3.27 & 3.15 & 5.62 & 3.53 & 3.47 & 5.31 & 3.17 & 2.97 & 5.89 & 3.9 & 4.65 & 9.88 & 1.89 & 2.34 & 5.16 & \textbf{1.22} & \textbf{1.79} & \textbf{3.45}\\ 
bowl& 2.61 & 2.3 & 4.55 & 2.66 & 2.35 & 3.97 & 2.46 & 2.16 & 4.84 & 4.35 & 5.0 & 14.59 & 1.77 & 1.97 & 3.9 & \textbf{1.05} & \textbf{1.32} & \textbf{2.4}\\ 
camera& 4.4 & 4.78 & 7.85 & 4.84 & 5.3 & 8.03 & 4.24 & 4.43 & 8.11 & 6.78 & 8.04 & 13.91 & 2.31 & 3.38 & 7.2 & \textbf{1.63} & \textbf{2.67} & \textbf{4.97}\\ 
can& 1.95 & 1.73 & 5.86 & 1.95 & 1.89 & 5.21 & 2.02 & 1.7 & 5.82 & 2.95 & 3.47 & 23.02 & 1.53 & 1.8 & 3.08 & \textbf{0.8} & \textbf{1.17} & \textbf{2.85}\\ 
cap& 6.07 & 5.98 & 11.49 & 7.21 & 7.14 & 10.94 & 4.68 & 4.23 & 9.17 & 14.11 & 14.86 & 28.23 & 3.29 & 4.87 & 13.02 & \textbf{1.4} & \textbf{2.74} & \textbf{8.35}\\ 
keyboard& 0.98 & 0.96 & 1.35 & 1.07 & 1.0 & 1.23 & 0.79 & 0.77 & 1.55 & 1.13 & 1.16 & 2.58 & 0.73 & 0.77 & 1.11 & \textbf{0.43} & \textbf{0.45} & \textbf{0.63}\\ 
dishwasher& 2.09 & 1.8 & 4.55 & 2.45 & 2.09 & 3.53 & 2.51 & 1.77 & 4.72 & 3.44 & 3.78 & 9.31 & 1.79 & 1.7 & 3.27 & \textbf{0.93} & \textbf{1.05} & \textbf{2.04}\\ 
earphone& 6.86 & 6.96 & 12.77 & 7.88 & 6.59 & 16.53 & 5.33 & 4.83 & 11.67 & 20.31 & 23.21 & 39.49 & 4.29 & \textbf{4.16} & \textbf{10.3} & \textbf{2.03} & 5.1 & 10.69\\
helmet& 4.86 & 5.04 & 8.86 & 6.15 & 6.41 & 9.16 & 4.89 & 4.86 & 8.73 & 8.78 & 10.07 & 21.2 & 3.06 & 4.38 & 10.27 & \textbf{1.86} & \textbf{3.3} & \textbf{6.96}\\ 
mailbox& 2.2 & 2.29 & 4.49 & 2.74 & 2.68 & 4.31 & 2.35 & 2.2 & 4.91 & 5.2 & 5.33 & 10.94 & 1.52 & 1.9 & 4.33 & \textbf{1.03} & \textbf{1.47} & \textbf{3.34}\\ 
microphone& 2.92 & 3.27 & 8.54 & 4.36 & 4.65 & 8.46 & 3.03 & 3.2 & 7.15 & 6.39 & 7.99 & 19.41 & 2.29 & 3.23 & 8.41 & \textbf{1.25} & \textbf{2.27} & \textbf{5.47}\\ 
microwaves& 2.29 & 2.12 & 5.17 & 2.59 & 2.35 & 4.47 & 2.67 & 2.12 & 5.41 & 3.89 & 4.08 & 9.01 & 1.74 & 1.81 & 3.82 & \textbf{1.01} & \textbf{1.18} & \textbf{2.14}\\ 
pillow& 2.07 & 2.11 & 3.73 & 2.09 & 2.16 & 3.54 & 2.08 & 2.05 & 4.01 & 4.15 & 4.29 & 12.01 & 1.43 & 1.69 & 3.43 & \textbf{0.92} & \textbf{1.24} & \textbf{2.39}\\ 
printer& 3.02 & 3.23 & 5.53 & 3.28 & 3.6 & 5.56 & 2.9 & 2.96 & 6.07 & 5.38 & 5.94 & 10.29 & 1.82 & 2.41 & 5.09 & \textbf{1.18} & \textbf{1.76} & \textbf{3.1}\\ 
remote& 0.89 & 0.92 & 1.85 & 0.95 & 1.08 & 1.58 & 0.89 & 0.89 & 2.28 & 1.51 & 1.75 & 6.0 & 0.82 & 1.02 & 1.29 & \textbf{0.44} & \textbf{0.58} & \textbf{0.78}\\ 
rocket& 1.28 & 1.09 & 2.0 & 1.39 & 1.22 & 2.01 & 1.14 & 0.96 & 2.03 & 1.84 & 1.51 & 4.01 & 0.97 & 0.79 & 1.6 & \textbf{0.39} & \textbf{0.72} & \textbf{1.39}\\ 
skateboard& 1.53 & 1.42 & 1.99 & 1.97 & 1.78 & 2.45 & 1.23 & 1.2 & 2.01 & 2.43 & 2.53 & 4.25 & 0.93 & 1.07 & 1.83 & \textbf{0.52} & \textbf{0.8} & \textbf{1.31}\\ 
tower& 2.25 & 2.25 & 4.74 & 2.37 & 2.4 & 4.35 & 2.2 & 2.17 & 5.47 & 3.38 & 4.15 & 13.11 & 1.35 & 1.8 & 3.85 & \textbf{0.82} & \textbf{1.35} & \textbf{2.48}\\ 
washer& 2.58 & 2.34 & 5.5 & 2.77 & 2.52 & 4.64 & 2.63 & 2.14 & 6.57 & 4.53 & 4.27 & 9.23 & 1.83 & 1.97 & 5.28 & \textbf{1.04} & \textbf{1.39} & \textbf{2.73}\\ 
\hline
mean& 2.79 & 2.77 & 5.49 & 3.22 & 3.13 & 5.43 & 2.65 & 2.46 & 5.52 & 5.37 & 5.95 & 13.55 & 1.84 & 2.23 & 4.95 & \textbf{1.05} & \textbf{1.67} & \textbf{3.45}\\ 

\bottomrule
\end{tabular}
\end{adjustbox}
\end{table*}

\section{Complexity Analysis}
Our method achieves the best performance on both our newly proposed diverse benchmarks and the existing benchmarks. We provide the detailed complexity analysis of our method in Table~\ref{tab:FLOPs}. We report the number of parameters and  theoretical computation cost (FLOPs) of our method and other five methods. We also provide the average Chamfer distances of all categories in ShapeNet-55 and unseen categories in ShapeNet34 as references. We can see our method achieves the best performance while using relatively low parameters and FLOPs among the methods in the table, which shows our method offers a decent trade-off between cost and performance.

\begin{table}[h]
\small
\caption{Complexity analysis. We report the the number of parameter (Params) and theoretical computation cost (FLOPs) of our method and five existing methods. We also provide the average Chamfer distances of all categories in ShapeNet-55 (CD$_{\rm 55}$) and unseen categories in ShapeNet34 (CD$_{\rm 34}$) as references.} 
\label{tab:FLOPs}
\centering
\setlength{\tabcolsep}{1.5mm}{
\begin{tabular}[\linewidth]{l | r r | r r }
\toprule
Models & Params & FLOPs & CD$_{\rm 55}$ &  CD$_{\rm 34}$\\
\hline
FoldingNet~\cite{FoldingNet}&2.30 M &27.58 G &3.12 & 3.62 \\
PCN~\cite{PCN}     & 5.04 M & 15.25 G & 2.66 & 3.85\\
TopNet~\cite{TopNet}    &5.76 M &6.72 G & 2.91 & 3.50\\
PFNet~\cite{PFNet}   & 73.05 M & 4.96 G & 5.22 & 8.16 \\
GRNet~\cite{GRNet}    & 73.15 M & 40.44 G & 1.97 & 2.99 \\
\hline
PoinTr  & 30.9 M & 10.41 G & 1.07 & 2.05 \\

\bottomrule[1pt]
\end{tabular}}
\end{table}

\section{Visualization of the Predicted Centers}
\begin{figure}[t]
  \centering
  \includegraphics[width = \linewidth]{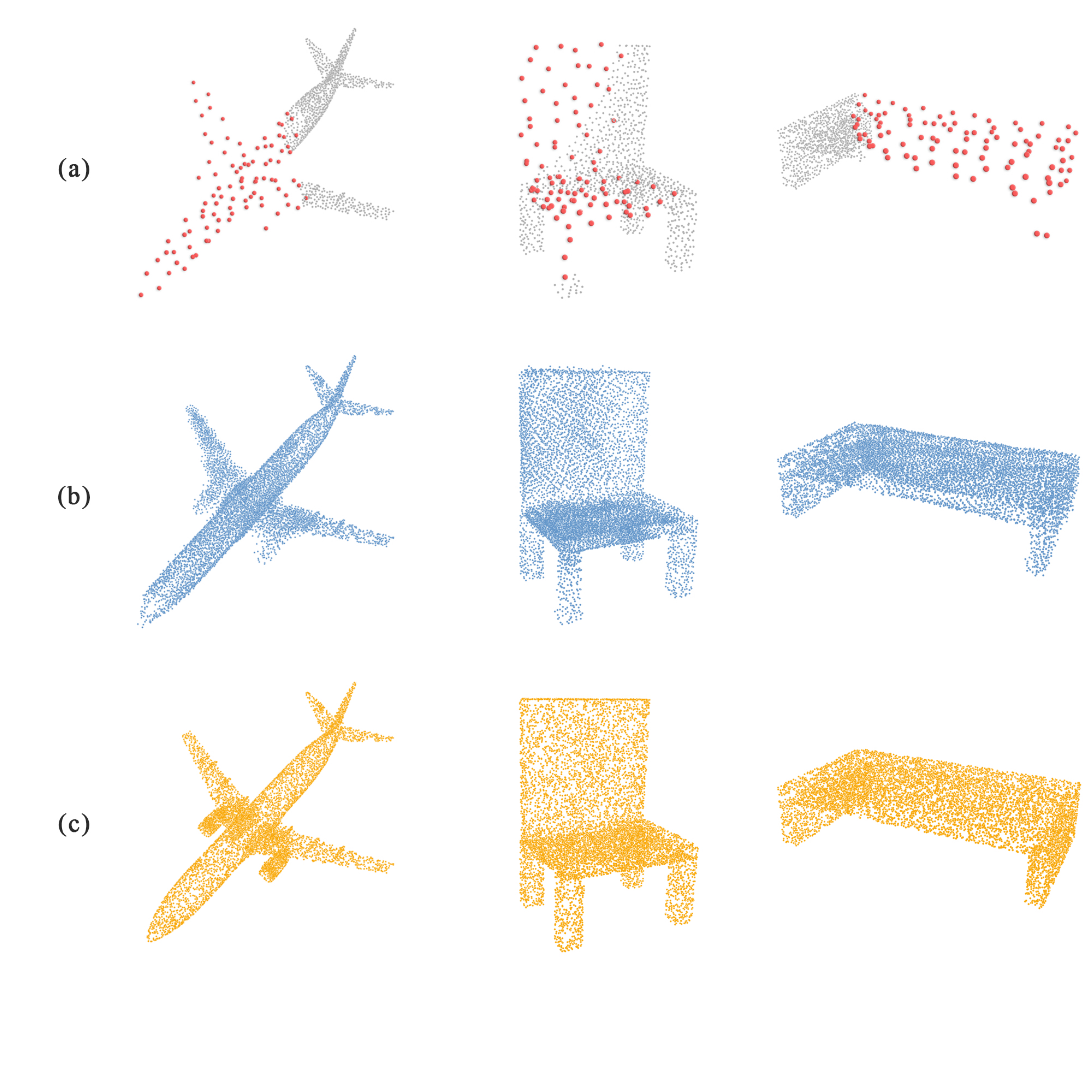}
  \caption{\small Visualization of predicted points proxies. In Line (a), we show the input partial point clouds and the predicted centers. Based on predicted point proxies, we can easily predicted the accurate point centers and then complete the point clouds, as shown in Line (b). We show the ground-truth point cloud in Line (c) for comparisons.}
  \label{fig:proxy}
\end{figure}

We visualize the local center prediction results on ShapeNet-55. We adopt a coarse-to-fine strategy to recover the point cloud. Our method starts with the prediction of local centers, then we can obtain the final results by adding the points around the centers. As shown in Figure~\ref{fig:proxy}, Line (a) shows the input partial point cloud and the predicted point centers. Line (b) is the predicted point clouds. We see the predicted point proxies can successfully represent the overall structure of the point cloud and the details then are added in the final predictions.

\section{Qualitative Results}
 \begin{figure*}[th]
  \centering
  \includegraphics[width = 0.9 \linewidth]{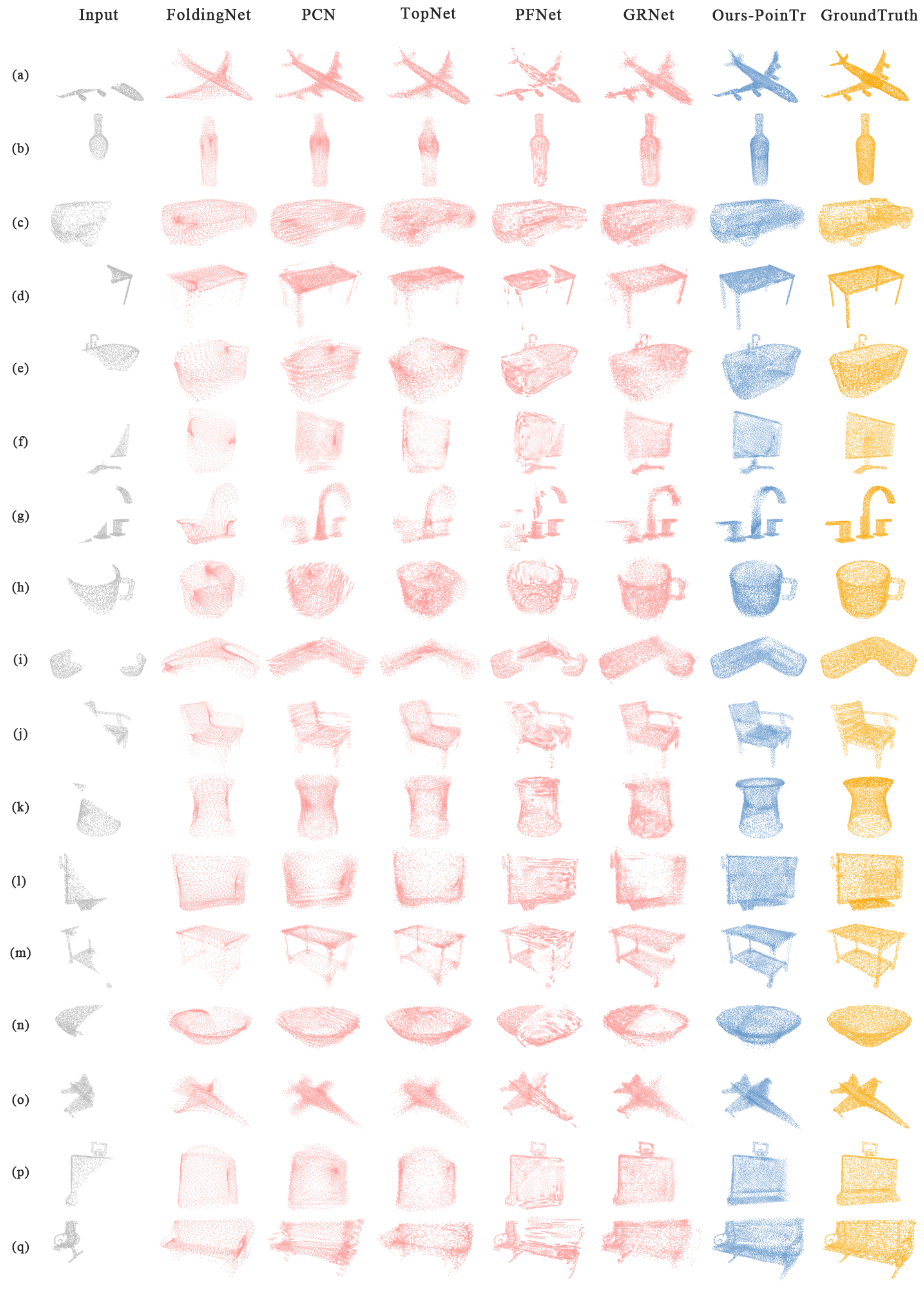}
  \caption{\small More qualitative results on ShapeNet-55.}
  \label{fig:supp_case}
\end{figure*}

In Figure~\ref{fig:supp_case}, we provide more qualitative results on ShapeNet-55. We see our results are much better than baseline methods visually.

\end{appendix}

{\small
\bibliographystyle{ieee_fullname}
\bibliography{egbib}
}
\end{document}